\documentclass[journal]{IEEEtran}

\usepackage{amsmath,amsfonts,bm}

\def\eqref#1{equation~\ref{#1}}

\def\1{\bm{1}}

\def\rva{{\mathbf{a}}}
\def\rvb{{\mathbf{b}}}

\def\rvh{{\mathbf{h}}}

\def\rvp{{\mathbf{p}}}

\def\rvx{{\mathbf{x}}}

\def\rvy{{\mathbf{y}}}

\def\erva{{\textnormal{a}}}

\def\ervu{{\textnormal{u}}}

\def\rmA{{\mathbf{A}}}

\def\rmW{{\mathbf{W}}}

\def\ermA{{\textnormal{A}}}

\def\vtheta{{\bm{\theta}}}

\def\vh{{\bm{h}}}

\def\vp{{\bm{p}}}

\def\vx{{\bm{x}}}
\def\vy{{\bm{y}}}
\def\vz{{\bm{z}}}

\def\mH{{\bm{H}}}
\def\mI{{\bm{I}}}

\DeclareMathAlphabet{\mathsfit}{\encodingdefault}{\sfdefault}{m}{sl}
\SetMathAlphabet{\mathsfit}{bold}{\encodingdefault}{\sfdefault}{bx}{n}

\newcommand{\E}{\mathbb{E}}

\newcommand{\KL}{D_{\mathrm{KL}}}
\newcommand{\Var}{\mathrm{Var}}

\newcommand{\Cov}{\mathrm{Cov}}
\newcommand{\Proj}{\mathrm{Proj}}
\newcommand{\Alg}{\mathrm{Alg}}

\DeclareMathOperator*{\argmax}{arg\,max}
\DeclareMathOperator*{\argmin}{arg\,min}

\DeclareMathOperator{\Tr}{Tr}

\def\comma{{ \text{ ,} }}
\def\period{{ \text{ .} }}

\def\cspace{{\mathbb{R}^{m}}}
\def\Cspace{{\mathbb{R}^{\bar{m}}}}
\def\Xspace{{\mathbb{R}^{n}}}
\def\xyspace{{\mathbb{R}^{n+d}}}
\def\yspace{{\mathbb{R}^{d}}}
\def\expdnn{{\mathbb{E}_{\substack{ (\rvx_t, \rvy) \sim \mu_t \\ \text{subject to (\ref{eq:fc-dnn-dynamics})}}}}}%

\def\transpose{{\mathsf{T}}}

\def\gradsample{{ g^{i}(\vtheta) }}
\def\gradmb{{ g^{mb}(\vtheta) }}
\def\gradfull{{ g^{}(\vtheta) }}
\def\gradmbt{{   g^{mb}(\vtheta_{t}) }}
\def\gradfullt{{ g^{}(\vtheta_{t}) }}
\def\diffusionmatrixx{{ \mathbf{\Sigma}_{\mathcal{D}} }}
\def\diffusionmatrix{{ \mathbf{\Sigma}_{\mathcal{D}}(\vtheta) }}
\def\diffusionmatrixt{{ \mathbf{\Sigma}_{\mathcal{D}}(\vtheta_{t}) }}
\def\diffusionmatrixthaft{{ \mathbf{\Sigma}^{\frac{1}{2}}_{\mathcal{D}}(\vtheta_{t}) }}
\def\diffusionmatrixthaftt{{ \mathbf{\Sigma}^{\frac{1}{2}}_{\mathcal{D}} }}
\def\diffusionmatrixthafttt{{ \tilde{\mathbf{\Sigma}}^{\frac{1}{2}}_{\mathcal{D}} }}
\def\diffusionmatrixttt{{ \tilde{\mathbf{\Sigma}}_{\mathcal{D}} }}
\def\diffusionmatrixbatch{{ \tilde{\mathbf{\Sigma}}_{\mathcal{B}} }}
\def\absB{{ \mid \mathcal{B} \mid }}
\def\absD{{ \mid \mathcal{D} \mid }}
\def\drift{{ b(\vtheta_{t}) }}
\def\diffusion{{ \sigma(\vtheta_{t}) }}
\def\dt{{ \mathrm{d} t }}
\def\ds{{ \mathrm{d} s }}
\def\dWt{{ \mathrm{d} \mathbf{W}_t }}
\def\dLat{{ \mathrm{d} L_t^{\alpha} }}
\def\tt{{ \vtheta_{t} }}

\def\pss{{ \rho^{\mathrm{ss}} }}
\def\psss{{ \rho^{\mathrm{ss}}(\vtheta) }}

\def\Dk{{ \mathrm{D}^k }}
\def\gmbupdate{{ \vtheta^{\prime} - \eta g^{mb}(\vtheta^{\prime}) }}
\def\gmbupdatee{{ \vtheta - \eta g^{mb}(\vtheta) }}
\def\deltaa{{ \delta\left\{ \vtheta - \left[\gmbupdate\right] \right\} }}
\def\thprime{{ \vtheta^{\prime} }}
\def\dthprime{{ \mathrm{d} \thprime }}
\def\dth{{ \mathrm{d} \vtheta }}
\def\expB{{   \mathbb{E}_{\mathcal{B}} }}
\def\expPss{{ \mathbb{E}_{\pss} }}
\def\smallint{ \int}
\def\smallsum{\begingroup\textstyle \sum \endgroup}

\def\expBof#1{ \expB \left[ {#1} \right]}
\def\expPssof#1{ \expPss \left[ {#1} \right]}
\def\Phigmd{{ \Phi \left(\gmbupdatee\right) }}

\def\itoarg{{\left( X_t, t \right) }}

\def\thetap{{ \vtheta^{\prime} }}
\def\LocalEntropy{{ - \log \int_{\thetap \in \cspace} \exp (-\Phi (\thetap)-\frac{\gamma}{2}\|\vtheta-\thetap\|_{2}^{2}) \mathrm{d} \thetap }}

\def\ExpOfXi#1{ \mathbb{E}_\xi \left[ {#1} \right]}

\def\fracK {{\frac{1}{k}}}

\def\metanext{{ \vtheta_{\text{adapt}}^{n+1} }}
\def\metacurr{{ \vtheta_{\text{adapt}}^{n}   }}

\usepackage{url}
\usepackage{hyperref}
\hypersetup{
    colorlinks,
    linkcolor={red!80!black},
    citecolor={red!80!black},
    urlcolor={blue!80!black}
}

\usepackage{enumitem}
\usepackage{lipsum}
\usepackage{blindtext}
\usepackage{amsthm}
\usepackage{cancel}
\usepackage{mathtools}
\usepackage{xcolor}
\usepackage{fancyhdr}

\usepackage{amsmath,amssymb}

\colorlet{darkgreen}{green!70!black}

\newtheorem{theorem}{Theorem}
\newtheorem{lemma}{Lemma}

\newcommand{\eq}[1]{{(#1)}}

\newcommand{\citet}[1]{{\cite{#1}}}
\newcommand{\citep}[1]{{\cite{#1}}}

\newcommand\numberthis{\addtocounter{equation}{1}\tag{\theequation}}

\usepackage{cite}

\ifCLASSINFOpdf
  \usepackage[pdftex]{graphicx}
\else
\fi

 \usepackage[caption=false,font=footnotesize]{subfig}

\begin{document}
\title{Deep Learning Theory Review: An Optimal Control and Dynamical Systems Perspective}

\author{Guan-Horng~Liu,~
        Evangelos~A.~Theodorou~%
\thanks{The authors are with the
Autonomous Control and Decision Systems (ACDS) Lab, School of Aerospace Engineering,
Georgia Institute of Technology, Atlanta, GA 30332, USA (e-mail:
ghliu@gatech.edu;\newline evangelos.theodorou@gatech.edu)}%
\thanks{
\copyright $\text{ }$ 2019 IEEE. Personal use of this material is permitted. Permission
from IEEE must be obtained for all other uses, in any current or future
media, including reprinting/republishing this material for advertising or
promotional purposes, creating new collective works, for resale or
redistribution to servers or lists, or reuse of any copyrighted
component of this work in other works.}
}


\maketitle

\begin{abstract}
Attempts from different disciplines to provide a fundamental understanding of deep learning have advanced rapidly in recent years,
yet
a unified framework remains relatively limited.
In this article, we provide one possible way to align existing branches of deep learning theory
through the lens of dynamical system and optimal control.
By viewing deep neural networks as discrete-time nonlinear dynamical systems,
we can analyze how information propagates through layers using mean field theory.
When optimization algorithms are further recast as controllers,
the ultimate goal of training processes can be formulated as an optimal control problem.
In addition, we can reveal convergence and generalization properties by studying the stochastic
dynamics of optimization algorithms.
This viewpoint features a wide range of theoretical study
from information bottleneck to statistical physics.
It also provides a principled way for hyper-parameter tuning
when optimal control theory is introduced.
Our framework fits nicely with supervised learning and can be extended to
other learning problems, such as Bayesian learning, adversarial training, and specific forms of meta learning,
without efforts.
The review aims to shed lights on the importance of dynamics and optimal control
when developing deep learning theory.
\end{abstract}


\IEEEpeerreviewmaketitle

\section {Introduction}
\IEEEPARstart{D}{eep} learning is one of the most rapidly developing  areas in modern artificial intelligence  with tremendous impact to different industries ranging from the areas of social media, health and biomedical engineering, robotics,  autonomy and aerospace systems.
 Featured with millions of parameters yet without much hand tuning or domain-specific knowledge, Deep Neural Networks (DNNs)
 match and sometimes exceed human-level performance in complex problems involving visual synthesizing \citep{krizhevsky2012imagenet}, language reasoning \citep{srivastava2012multimodal}, and long-horizon consequential planning \citep{silver2016mastering}.
The remarkable practical successes, however, do not come as a free lunch.
	Algorithms for training  DNNs are extremely data-hungry.
	While insufficient data can readily lead to memorizing irrelevant features \citep{geirhos2018imagenet},
    data imbalance may cause severe effects such as
    {imposing improper priors implicitly \citep{wang2016training}.}
	{ Although automating tuning  for millions of parameters alleviates inductive biases from traditional engineering, it comes with the drawback of
     making  interpretation and analysis  difficult}.
	Furthermore,
	DNN is known for being
	vulnerable to small adversarial perturbations \citep{goodfellow2014explaining}.
	In fact, researchers have struggled to improve its robustness beyond infinite norm ball attacks \citep{athalye2018robustness}.
Finally, while the deep learning community has developed recipes related to the choice of the underlying organization of a DNN, the process of the  overall architectural design lacks solid theoretical understanding and remains a fairly ad-hock process.

The aforementioned issues  highlight the need towards  the development of  a theory for deep learning which  will provide a scientific methodology to design  DNNs architectures,  robustify their performance against external attacks/disturbances, and enable the development  the corresponding training algorithms.
Given this need, our objective in this paper is to review and present in a systematic way work towards the discovery of deep learning theory.
This work relies on concepts drawn primarily from the areas of dynamical systems and  optimal control theory, and its connections to information theory and statistical physics.

Theoretical understanding of DNN training from previous works has roughly followed two streams: deep latent representation and stochastic optimization.
On the topic of deep representations, the composition of affine functions,
with element-wise nonlinear activations,
plays a crucial role in automatic feature extraction \citep{simonyan2013deep}
{by constructing a chain of differentiable process.}
{An increase in the depth of a NN architecture has the effect of increasing its expressiveness exponentially \citep{poole2016exponential}.}
This naturally yields a highly over-parametrized model,
whose loss landscape is known for a proliferation of local minima and saddle points
\citep{dauphin2014identifying}.
However, the over-fitting phenomena, suggested by the bias-variance analysis, has not been observed during DNN training \citep{zhang2016understanding}.
In practice, DNN often generalizes remarkably well on unseen data when initialized properly \citep{sutskever2013importance}.

Generalization of highly over-parametrized models cannot be properly explained without
considering stochastic optimization algorithms.
Training DNN is a non-convex optimization problem.
	Due to its high dimensionality, most practically-used algorithms utilize first-order derivatives with aids of adaptation and momentum mechanisms.
	In fact, even a true gradient can be too expensive to compute on the fly;
	therefore only an unbiased estimation is applied at each update.
{Despite these  approximations that are typically used to enable applicability,
first-order stochastic optimization is surprisingly robust and
algorithmically stable \citep{hardt2015train}.}
	The stochasticity stemmed from estimating gradients is widely believed to perform implicit regularization \citep{chaudhari2018stochastic},
	guiding parameters towards flat plateaus with lower generalization errors \citep{keskar2016large}.
	First-order methods are also proven more efficient to escape from saddle points \citep{lee2017first},
	whose number grows exponentially with model dimensionality \citep{dauphin2014identifying}.
Research along this line
provides
{a fundamental understanding} on training dynamics and convergence property,
despite the analysis is seldom connected to the deep representation viewpoint.

 {How do the two threads of deep latent organization and stochastic optimization interplay with each other  and what are the underlying theoretical connections? These are questions that have not been well-explored and are essential towards the development of a theory for Deep Learning.}
Indeed, study of stochastic optimization dynamics often treats DNN merely as a black-box.
This may be insufficient to describe the whole picture.
	{When using back-propagation \citep{lecun1990handwritten} to obtain first-order derivatives,
	the backward dynamics, characterized by the compositional structure, rescales the propagation made by optimization algorithms,
	which in return leads to different representation at each layer.}
Frameworks that are able to mathematically characterize these compounding effects will provide more nuanced statements.
	One of such attempts has been information bottleneck theory \citep{shwartz2017opening},
	which describes the dynamics of stochastic optimization using information theory, and connects it to optimal representation via the bottleneck principle.
	Another promising branch from Du \textit{et al.} \cite{du2018gradientb,du2018gradienta} showed that
	the specific representation, i.e. the Gram matrix, incurred from gradient descent (GD)
	characterizes the dynamics of the prediction space and can be used to prove global optimality.
	These arguments, however, have been limited to either certain architectures \citep{saxe2018information}
	or noise-free optimization\footnote{
		We note that global optimality for stochastic gradient descent has been recently proven
		in \cite{zou2018stochastic,allen2018convergence},
		yet their convergence theories rely on certain assumptions on the data set in order to have the Frobenius norm of the (stochastic) gradient lower-bounded.
		This is in contract the least eigenvalue of the prediction dynamics in \cite{du2018gradientb,du2018gradienta},
		which is more related to the dynamical analysis in this review.
	}.

{ In this review, we provide a dynamical systems and optimal control perceptive to DNNs in an effort to systematize the alternative approaches and methodologies.}
This allows us to pose and answer the following questions:
(i) at which state should the training trajectory start, i.e. how should we initialize the weights or hyper-parameters,
(ii) through which path, in a distribution sense, may the trajectory traverse,
i.e. can we give any prediction of training dynamics on average,
(iii) to which fixed point, if exists, may the trajectory converge, and finally
(iv) the stability and generalization property at that fixed point.
In the context of deep learning, these can be done
by recasting DNNs and optimization algorithms as (stochastic) dynamical systems.
Advanced tools from signal processing, mean-field theory, and stochastic calculus can then be applied to better reveal the training properties.
We can also formulate an optimal control problem upon the derived dynamics to provide principled guidance for architecture and algorithm design.
The dynamical and control viewpoints fit naturally with supervised learning and can readily extend to
other learning schemes, such as Bayesian learning, adversarial training, and specific forms of meta learning,
This highlights the potential to provide more theoretical insights.

The article is organized as follows.
In Sec. \ref{sec:related-work}, we will go over recent works related to the dynamical viewpoint.
Sec. \ref{sec:dnn-dynamics} and \ref{sec:sgd} demonstrate how we can recast DNNs and stochastic optimizations as dynamical systems,
then apply control theory to them.
In Sec. \ref{sec:beyond-sl}, we extend our analysis to other learning problems.
Finally, we conclude our work and discuss some future directions in Sec. \ref{sec:diss-conclusion}.

\textbf{Notation:}
We will denote
$\vh_l$ and $\vx_l$ as the (pre-)activation at layer $l$ ($\vx_0 \equiv \vx$ for simplicity).
Mapping to the next layer obeys
$\vh_l = \mathcal{W}(\vx_l; \vtheta_l)$ and $\vx_{l+1} = \phi(\vh_l)$, where
$\phi$ is a nonlinear activation function and
$\mathcal{W}$ is an affine transform parametrized by $\vtheta_l \in \cspace$.
The full parameter space across all layers is denoted $\vtheta \equiv \{\vtheta_l\}_{l=0}^{L-1} \in \Cspace$.
Given a data set
$\mathcal{D} := \{ (\vx^{(i)}, \vy^{(i)}) \}_i$,
where $\vx \in \Xspace$ and $\vy \in \yspace$,
we aim to minimize a loss $\Phi(\cdot)$, or equivalently the cost
$J(\cdot)$ from the control viewpoint.
The element of the vector/matrix are respectively denoted as
$\rva^{(i)} \equiv \erva_i$ and
$\rmA^{(i,j)} \equiv \ermA_{(i,j)}$.
We will follow the convention
$\langle \cdot , \cdot \rangle$
to denote the inner product of two vectors,
with
$\langle f(\cdot) , g(\cdot) \rangle_{\mu} := \int_{\Omega} f(w)^\transpose g(w) \mathrm{d}\mu(w)$
as its generalization to the inner product of two functions weighted by a probability measure.
We will always use the subscript $t$ to denote the dynamics.
Depending on the context, it can either mean propagation through DNN (Sec. \ref{sec:dnn-dynamics}) or training iterations (Sec. \ref{sec:sgd}).

\section {Related Work}
\label{sec:related-work}

\subsection{Mean Field Approximation \& Gaussian Process}
Mean field theory allows us to describe distributions of activations and pre-activations over an ensemble of untrained DNNs in an analytic form.
The connection was adapted in \citet{poole2016exponential} to study how signals propagate through layers at the initialization stage.
It implies an existence of an \textit{order-to-chaos} transition as a function of parameter statistics.
While networks in the phase of \textit{order} suffer from saturated information and vanished gradients, %
in the \textit{chaotic}  regime  expressions of networks grow exponentially with depth, and exploded gradients can be observed.
This phase transition diagram, formally characterized in \cite{schoenholz2016deep},
provides a necessary condition towards  network \textit{trainability}
and determines an upper bound on the number of layers allowable for information to pass through.
This  analysis has been successfully applied to most commonly-used architectures for critical initialization, including FCN, CNN, RNN, LSTM, ResNet
\citep{schoenholz2016deep,chen2018dynamical,gilboa2019dynamical,xiao2018dynamical,yang2017mean}.
In addition, it can also be used to provide geometric interpretation by estimating statistical properties of Fisher information \citep{karakida2018universal,pennington2018emergence}.
It is worth noticing that all aforementioned works require the limit of layer width and i.i.d. weight priors
in order to  utilize the Gaussian approximation.
Indeed, the equivalence between DNN and Gaussian process has long been known for single-layer neural networks
\citep{williams1997computing} and extended to deeper architectures recently \citep{matthews2018gaussian,garriga2018deep}.
The resulting Bayesian viewpoint enables uncertainty estimation and accuracy prediction at test time
\citep{lee2017deep}.

\subsection{Implicit Bias in Stochastic Gradient Descent (SGD)}
There has been commensurate interest in studying the implicit bias stemmed from stochastic optimization.
Even without stochasticity, vanilla GD algorithms are implicitly regulated as they converge to max-margin solutions
for both linear predictors
\citep{soudry2018implicit,gunasekar2018characterizing}
and over-parametrized models, e.g. multi-layers linear networks \citep{gunasekar2018implicit,ji2018gradient}
and shallow neural networks with nonlinear activations \citep{allen2018learning}.
When the stochasticity is introduced, a different regularization effect has been observed \citep{wu2018sgd}.
This implicit regularization plays a key role in explaining why DNNs generalize well despite {being over-parametrized}
\citep{zhang2016understanding,neyshabur2017geometry}.
Essentially, {stochasticity pushes} the training dynamics away from sharp local minima \citep{zhu2018anisotropic}
and instead guides it towards flat plateaus with lower generalization errors \citep{keskar2016large}.
An alternative view suggests a convolved, i.e. smoothening, effect on the loss function throughout training \citep{kleinberg2018alternative}.
The recent work from Chaudhari \textit{et al.} \cite{chaudhari2018stochastic}
provides mathematical intuitions by showing that SGD performs variational inference under certain approximations.
Algorithmically, Chaudhari \textit{et al.}
proposed a surrogate loss that explicitly biases SGD dynamics towards flat local minima \cite{chaudhari2016entropy,chaudhari2018deep}.
The corresponding algorithm relates closely to stochastic gradient Langevin dynamics,
a computationally efficient sampling technique originated from Markov Chain Monte Carlo (MCMC) for large-scale problems
\citep{teh2016consistency,li2016preconditioned}.

\subsection{Information Theory \& Statistical Physics}
Research along this direction studies the dynamics of Markovian stochastic process
and its effect on the deep representation at an ensemble level.
For instance, the Information Bottleneck (IB) theory \citep{tishby2015deep,shwartz2017opening}
studies the training dynamics on the Information Plane described by the mutual information of layer-wise activations.
{{} The principle of information bottleneck mathematically characterizes the phase transition from memorization to generalization,
mimicking the critical learning periods in biological neural systems \citep{achille2018critical}.
Applying the same information Lagrangian to DNN's weights has revealed intriguing properties of the deep representation, such as invariance, disentanglement and generalization \citep{achille2018emergence,DBLP:journals/corr/abs-1905-12213},
despite a recent debate in \cite{saxe2018information} arguing the inability  of the findings in references \citep{tishby2015deep,shwartz2017opening} to generalize beyond certain architectures.
In \cite{valle-perez2018deep}, similar statements on the implicit bias has been drawn from the Algorithmic Information Theory (AIT), suggesting the parameter-function map of DNNs is exponentially biased towards simple functions.}
The information theoretic viewpoint is closely related to statistical physics.
In \cite{goldt2017stochastic,goldt2017thermodynamic},
an upper bound mimicking the second law of stochastic thermodynamics was derived for single-layer networks on binary classification tasks.
In short, generalization of a network to unseen datum is bounded
by the summation of the Shannon entropy of its weights and { a term that captures the total heat dissipated during training.}
The concept of \textit{learning efficiency} was introduced as an alternative metric to compare algorithmic performance.
Additionally, Yaida \textit{et al.} \cite{yaida2018fluctuation} derived a discrete-time master equation at stationary equilibrium and linked it to the fluctuation-dissipation theorem in statistical mechanics.

\subsection{Dynamics and Optimal Control Theory}
The dynamical perspective has received considerable attention recently as it brings new insights to deep architectures and training processes.
	For instance, viewing DNNs as a discretization of continuous-time dynamical systems is proposed in \cite{weinan2017proposal}.
	From such, the propagating rule in the deep residual network \citep{he2016deep},
	$\vx_{t+1} = \vx_t + f(\vx_t, \vtheta_t)$,
	can be thought of as an one-step discretization of the forward Euler scheme on an ordinary differential equation (ODE), $\dot{\vx_t} = f(\vx_t, \vtheta_t)$.
	This interpretation has been leveraged to improve residual blocks in the sense that it achieves more effective numerical approximation \citep{lu2017beyond}.
	{{} In the continuum limit of depth, the flow representation of DNNs has made the transport analysis with Wasserstein geometry possible \citep{JMLR:v20:16-243}.
	Algorithmically,
	efficient computational methods have been developed in \cite{chen2018neural,chen2019cheapdiffopt} to allow parameterization of (stochastic) continuous-time dynamics (e.g. derivative of latent variables) directly with DNNs.}
When the analogy between optimization algorithms and controllers is further drawn \citep{hu2017control}, standard supervised learning can be recast as a mean-field optimal control problem \citep{han2018mean}.
This is particularly beneficial since it enables new training algorithms inspired from optimal control literature \citep{li2017maximum,li2018optimal,zhang2019you}.

Similar analysis can be applied to SGD by viewing it as a stochastic dynamical system.
In fact,
most previous discussions on implicit bias formulate SGD as stochastic Langevin dynamics \citep{pavliotis2014stochastic}.
Other stochastic modeling, such as L\`{e}vy process, has been recently proposed  \citep{simsekli2019tail}.
In parallel, stability analysis of the Gram matrix dynamics induced by DNN reveals global optimality of GD algorithms
\citep{du2018gradienta,du2018gradientb}.
Applying optimal control theory to SGD dynamics results in optimal adaptive strategies for tuning hyper-parameters,
such as the learning rate, momentum, and batch size
\citep{li2017stochastic,an2018stochastic}.

\section {Deep Neural Network as a Dynamical System}
\label{sec:dnn-dynamics}

As mentioned in Sec. \ref{sec:related-work},
DNNs can be interpreted as finite-horizon nonlinear dynamical systems by viewing each layer as a distinct time step.
In Sec \ref{sec:info-prop-dnn}, we will discuss how to explore this connection to analyze information propagation inside DNN.
The formalism establishes the foundation of recent works \citep{du2018gradientb,schoenholz2016deep,xiao2018dynamical}, and we will discuss its implications in Sec. \ref{sec:dnn-dynamics-impl}.
In Sec \ref{sec:dnn-oc}, we will draw the connection between optimization algorithms and controllers, leading to an optimal control formulation of DNN training characterized by mean-field theory.
Hereafter we will focus on fully-connected (FC) layers and leave remarks for other architectures,
e.g. convolution layers and residual connections, in Appendix \ref{app:dnn-dynamcs-other-archi}.

\subsection {Information Propagation inside DNN}
\label{sec:info-prop-dnn}

\begin{figure*}[!t]
\centering
\subfloat[
The phase diagram of $\max(\chi_{q^*}, \chi_{c^*})$ as the function of $\sigma_w$ and $\sigma_b$.
The solid line represents the critical initialization where information inside DNN is neither saturated, as in the ordered phase, nor exploding, as in the chaotic phase.
]{\includegraphics[width=0.8\columnwidth]{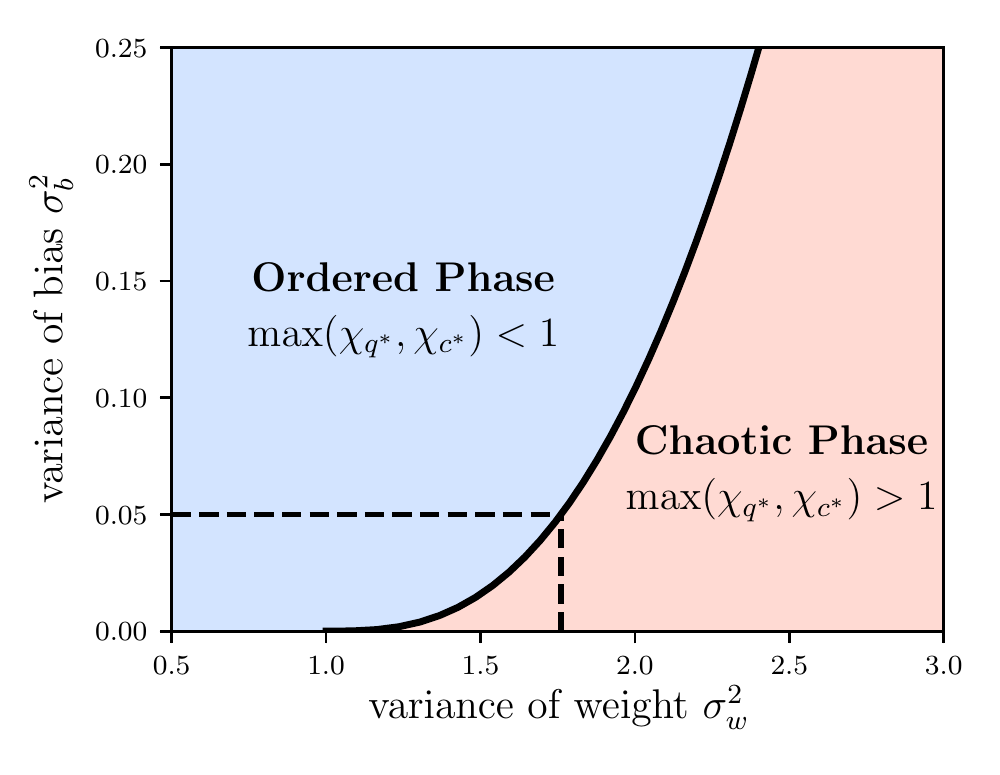}%
\label{fig:phase-diagram}}
\hfil
\subfloat[
Prediction of the network trainability given its depth and $\sigma_w^2$, with $\sigma_b^2$ fixed to $0.05$.
The color bar represents the training accuracy on MINST after $200$ training steps using SGD.
It is obvious that the boundary at which networks become un-trainable
aligns with the theoretical depth scale, denoted white dashed line, up to a constant ($\sim 4.5\xi_{c^*}$ in this case).
Also, notice that the peak around $\sigma_w^2 = 1.75$ is precisely
predicted by the critical line in Fig. \ref{fig:phase-diagram} for $\sigma_b^2 = 0.05$.
]{\includegraphics[width=0.8\columnwidth]{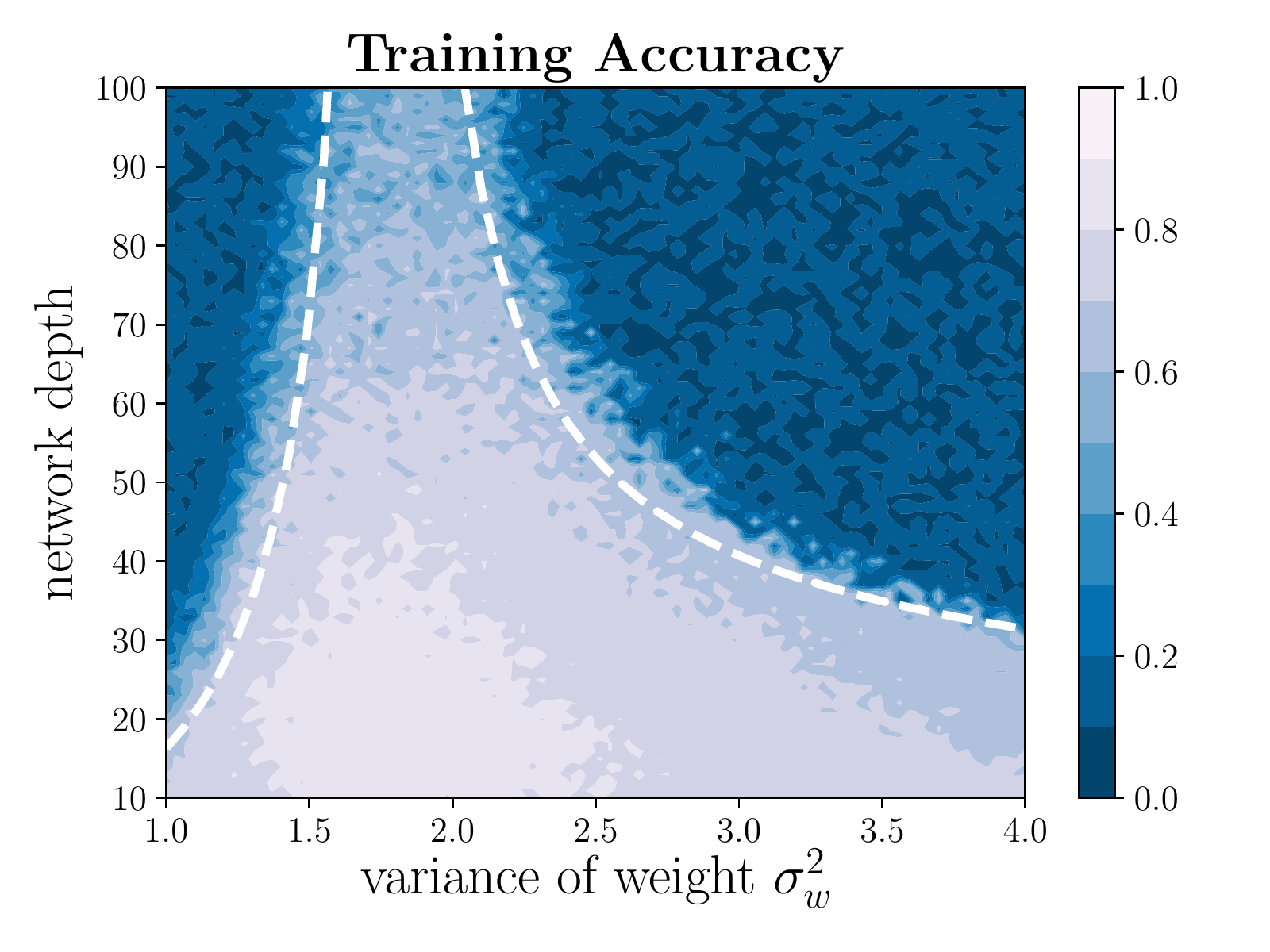}%
\label{fig:depth-scale}}
\caption{Reproduced results\protect\footnotemark
$\text{ }$from \cite{schoenholz2016deep} for random FC-DNNs with $\phi = \tanh$.}
\label{fig_sim}
\end{figure*}

Recall the dynamics of a FC-DNN at time step $t$, i.e. at layer $t$-th and
suppose its weights and biases are initialized by drawing i.i.d. from two zero-mean Gaussians, i.e.
\begin{equation}
\begin{split}
    \rvh_t &= \mathcal{W}_{\text{ FC}}(\rvx_t; \vtheta_t)
           := \rmW_t \rvx_t + \rvb_t \text{, where} \\
           \rmW_t^{(i,j)} &\sim \mathcal{N}(0, \sigma_{w}^2/N_{t}) \quad \rvb_t^{(i)} \sim \mathcal{N}(0, \sigma_{b}^2) \period \label{eq:fc-dnn-dynamics}
\end{split}
\end{equation}
We denote $\vtheta_t \equiv (\rmW_t, \rvb_t)$.
$\sigma_w^2$ and $\sigma_b^2$ respectively represent the variance of the weights and biases, and
$N_t$ is the dimension of the pre-activation at time $t$.
Central limit theorem (CLT) implies in the limit of large layer widths, $N_t \gg 1$,
the distribution of pre-activation elements, $\rvh_t^{(i)}$, also converges to a Gaussian.
It is straightforward to see the distribution also has zero mean
and its variance can be estimated by matching the second moment of the empirical distribution of $\rvh_t^{(i)}$ across all $N_t$,
\begin{align}
\label{eq:qt_def}
    q_t := \frac{1}{N_t} \sum_{i=0}^{N_t}(\rvh_t^{(i)})^2
    \period
\end{align}
$q_t$ can be viewed as the normalized squared length of the pre-activation, and
we will use it as the statistical quantity of the information signal.
The dynamics of $q_t$, when propagating from time step $t$ to $t+1$, takes a nonlinear form
\begin{align}
    q_{t+1}
    =    \sigma_w^2 \E_{\rvh_t^{(i)} \sim \mathcal{N}(0, q_t)}\left[\phi^2(\rvh_t^{(i)})\right] + \sigma_b^2 \comma
\label{eq:qt_mapping}
\end{align}
with the initial condition given by $q_0 = \frac{1}{N_0} \vx_0 \cdot \vx_0$.
Notice that
despite starting the derivation from random neural networks,
the mapping in \eq{\ref{eq:qt_mapping}}
admits a deterministic process, depending only on $\sigma_w$, $\sigma_b$, and $ \phi(\cdot)$.
We highlight this determinism as the benefit gained by mean-field approximation.
Schoenholz \textit{et al.} \cite{schoenholz2016deep} showed that for any bounded $\phi$ and finite value of $\sigma_w$ and $\sigma_b$,
there exists a fixed point, $q^* := \lim_{t \to \infty } q_t$, regardless of the initial state $q_0$.

Similarly, for a pair of input $(\vx^{(\alpha)}, \vx^{(\beta)})$ we can derive the following recurrence relation
\begin{align}
    q_{t+1}^{(\alpha, \beta)}
    = \sigma_w^2 \E_{(\rvh_t^{(\alpha)}, \rvh_t^{(\beta)})^\transpose \sim \mathcal{N}(\mathbf{0}, \mathbf{\Sigma}_{t}^{})}
                    [\phi(\rvh_t^{(\alpha)})\phi(&\rvh_t^{(\beta)})] + \sigma_b^2  \label{eq:qt_mapping2} \\
    \text{where } \quad \mathbf{\Sigma}_{t}^{} =
    \begin{pmatrix*}[l]
        q_t^\alpha & q_{t}^{(\alpha, \beta)} \\
        q_{t}^{(\alpha, \beta)} & q_t^\beta
    \end{pmatrix*}& \label{eq:qt_matrix}
\end{align}
is the covariance matrix at time $t$
and the initial condition is given by
$q_0^{(\alpha, \beta)} = \frac{1}{N_0} \vx_0^{(\alpha)} \cdot \vx_0^{(\beta)}$.
Under the same conditions for $q^*$ to exist,
we also have the fixed points for the covariance and correlation, respectively denoted
$q^{*}_{(\alpha, \beta)}$ and
$c^* = q^{*}_{(\alpha, \beta)} / \sqrt{q^{*}_{\alpha} q^{*}_{\beta}} = q^{*}_{(\alpha, \beta)} / q^{*}$.
The element of the covariance matrix at its fixed point $\mathbf{\Sigma}^{*}$  hence takes a compact form
\begin{align}
    {\Sigma}^{*}_{(\alpha, \beta)} = q^{*}\left(\delta_{(\alpha, \beta)} + (1  - \delta_{(\alpha, \beta)})c^*\right)\comma
\end{align}
where $\delta_{(a, b)}$ is Kronecker delta.
It is easy to verify that
$\mathbf{\Sigma}^{*}$
has $q^*$ for the diagonal entries and $q^*c^*$ for the off-diagonal ones.
In short,
when the mean field theory is applied to approximate distributions of activations and pre-activations,
the statistics of the distribution follows a deterministic dynamic as propagating through layers.
This statistic  can be treated as the information signal and from there the dynamical system analysis can be applied, as we will show in the next subsection.

\footnotetext{Code is available in \url{https://github.com/ghliu/mean-field-fcdnn}.}

\subsection {Stability Analysis \& Implications}
\label{sec:dnn-dynamics-impl}

Traditional stability analysis of dynamical systems often involves computing the Jacobian at the fixed points.
The Jacobian matrix describes the rate of change of system output when small disturbance is injected to input.
In this vein, we can define the residual system, $\bm{\epsilon}_t := \bm{\Sigma}_t - \bm{\Sigma}^{*}$, and first-order expand\footnote{We refer readers to the Supplementary Material in \citet{schoenholz2016deep} for a complete treatment.} it at $\bm{\Sigma}^{*}$.
The independent evolutions of the two signal quantities, as shown in \eq{\ref{eq:qt_mapping}} and (\ref{eq:qt_mapping2}),
already hint that the Jacobian can be decoupled into two sub-systems.
Their eigenvalues are given by,
\begin{align}
    \chi_{q^*} &= \sigma_w^2 \E_{\rvh \sim \mathcal{N}(0, \bm{\Sigma}^{*})}[\phi^{\prime\prime}(\rvh^{(i)})\phi(\rvh^{(i)}) + \phi^{\prime}(\rvh^{(i)})^2] \label{eq:chi_q} \\
    \chi_{c^*} &= \sigma_w^2 \E_{\rvh \sim \mathcal{N}(0, \bm{\Sigma}^{*})}[\phi^{\prime}(\rvh^{(i)})\phi^{\prime}(\rvh^{(j)})] \quad \rvh^{(i)} \neq \rvh^{(j)} \period
\end{align}
This eigen-decomposition suggests the information traverses through layers in the diagonal and off-diagonal eigenspace.
A fixed point is stable if and only if both $\chi_{q^*}$ and $\chi_{c^*}$ are less than $1$.
In fact, the logarithms of $\chi_{q^*}$ and $\chi_{c^*}$ relate to the well-known Lyapunov exponents in the dynamical system theory, i.e.
\begin{align}
\left|q_{t}-q^{*}\right| \sim e^{-t / \xi_{q^*}} \qquad &\xi_{q^*}^{-1} = - \log \chi_{q^*}  \quad \text{ and } \\
\left|c_{t}-c^{*}\right| \sim e^{-t / \xi_{c^*}} \qquad &\xi_{c^*}^{-1} = - \log \chi_{c^*}  \quad \comma \label{eq:xi_c}
\end{align}
where $c_t$ denotes the dynamics of the correlation.

The dynamical system analysis in   (\ref{eq:chi_q}-\ref{eq:xi_c}) has several important implications.
Recall again that given a DNN, its propagation rule depends only on $\sigma_w$ and $\sigma_b$.
We can therefore construct a phase diagram with $\sigma_w$ and $\sigma_b$ as axes and
define a critical line that separates the ordered phase,
in which all eigenvalues are less than $1$ to stabilize fixed points,
and the chaotic phase, in which either $\chi_{q^*}$ or $\chi_{c^*}$ exceeds $1$,
leading to divergence.
An example of the phase diagram for FC-DNNs is shown in Fig. \ref{fig:phase-diagram}.
    Networks initialized in the ordered phase may suffer from saturated information if the depth is sufficiently large.
    They become un-trainable since neither forward nor backward propagation is able to penetrate to the destination layer.
        Fig. \ref{fig:depth-scale} gives an illustration of how $\xi_{q^*}$ and $\xi_{c^*}$,
        named \textit{depth scale} \citep{schoenholz2016deep},
        predict the trainability of random DNNs.
    On the other hand, networks initialized along the critical line remain marginal stable, and
    information is able to propagate through an arbitrary depth without saturating or exploding.
    It is particularly interesting to note that while the traditional study on dynamical systems focuses on stability conditions,
    the dynamics inside DNN instead requires a form of transient chaos.

The discussion of the aforementioned critical initialization,
despite being crucial in the success of DNN training \citep{glorot2010understanding},
may seem limited
since once the training process begins, the i.i.d. assumptions in order to construct \eq{\ref{eq:qt_def}}, and all those derivations afterward, no longer hold.
Fortunately, it has been empirically observed that when DNN is sufficiently over-parameterized,
its weight will be close to the random initialization over training iterations \citep{li2018learning}.
{
In other words, under certain assumptions,
the statistical property derived at initialization can be preserved throughout training.
This has a strong implication
as it can be leveraged to prove
the global convergence and optimality of GD \citep{du2018gradienta,du2018gradientb}.
Below we provide the proof sketch and demonstrate how the deep information is brought into the analysis.
}

{
Recalling the information defined in   (\ref{eq:qt_mapping2}-\ref{eq:qt_matrix}) for a pair of input,
we can thereby construct a \textit{Gram matrix}
$\mathbf{K}_t \in \mathbb{R}^{\absD \times \absD}$,
where $|\mathcal{D}|$ is the size of the dataset and $t \in \{ 0,1,\cdots,T-1 \}$.
The element of $\mathbf{K}_t$ represents the information quantity between the data points with the corresponding indices,
i.e. $\mathbf{K}_t^{(i,j)} := q_t^{(i,j)}$.
The Gram matrix
can be viewed as a deep representation of a matrix form induced by the DNN architecture and dataset at random initialization.
{{}
The same matrix has been derived in several concurrent works, namely the Neural Tangent Kernel (NTK) in \cite{jacot2018neural}.}

Now,
consider a mean squared loss, $\frac{1}{2} \lVert \mathbf{y} - \mathbf{u} \rVert_{2}^2$,
where $\mathbf{y},\mathbf{u} \in \mathbb{R}^{\absD}$ and each element
$\mathbf{u}^{(i)} := \mathbf{1}^\transpose \vx^{(i)}_T / \lVert\mathbf{1}\rVert_2$
denotes the scalar prediction of each data point $i \in \mathcal{D}$.
The dynamics of the prediction error governed by the GD algorithm takes an analytical form \citep{du2018gradienta} written as
\begin{equation}
\begin{split}
    \mathbf{y}-\mathbf{u}(k+1) \approx& (\mathbf{I}-\eta \mathbf{G}(k))(\mathbf{y}-\mathbf{u}(k)) \comma \\
     \text{where }
     \mathbf{G}^{(i,j)}(k) :=& \langle \frac{\partial \ervu_i(k)}{\partial \theta_{T-1}(k)}
      ,\frac{\partial \ervu_j(k)}{\partial \theta_{T-1}(k)} \rangle
      \label{eq:gram-dynamics} \period
\end{split}
\end{equation}
$k$ and $\eta$ denote the iteration process and learning rate.
Equation \eq{\ref{eq:gram-dynamics}} indicates a linear dynamics characterized by the matrix $\mathbf{G}$,
whose element at initialization is related to the one of $\mathbf{K}$ by\footnote{
we set $\sigma_{b}^2$ to $0$ in \eq{\ref{eq:KGrelation}} to match the formulation used in \citep{du2018gradientb, du2018gradienta}.}
\begin{equation}
\begin{split}
     \mathbf{G}^{(i,j)}(0) &= \frac{1}{\sigma_{w}^2}
     \mathbf{K}_{T-1}^{(i,j)} \E_{\rvh \sim \mathcal{N}(0, \bm{\Sigma}_{T-1})}[\phi^{\prime}(\rvh^{(i)})\phi^{\prime}(\rvh^{(j)})] \\
     &=: \mathbf{K}_{T}^{(i,j)} \period
     \label{eq:KGrelation}
\end{split}
\end{equation}
When the width is sufficiently large, $\mathbf{G}(k)$ will be close to $\mathbf{K}_T$
(precisely $\lVert \mathbf{G}(k) - \mathbf{K}_T \rVert_2 $ is bounded)
for all iterations $k \geq 0$.
This, together with the least eigenvalue of $\mathbf{K}_T$ being lower-bounded for non-degenerate dataset, concludes the linear convergence to the global minimum.

}

\subsection {Training DNN with Optimal Control}
\label{sec:dnn-oc}

To frame optimization algorithms and training processes into the dynamical system viewpoint,
one intuitive way
is to interpret optimization algorithms as controllers.
As we will show in Sec. \ref{sec:mf-ocp},
this can be achieved without loss of generality and naturally yields an optimal control formalism of the deep learning training process.
Such a connection is useful since the optimality conditions of the former problem are well studied and characterized by
the Pontryagin's Minimum Principle (PMP) and
the Hamilton-Jacobi-Bellman (HJB) equation,
which we will introduce in Sec. \ref{sec:mf-pmp} and \ref{sec:mf-hjb}, respectively.
The fact that back-propagation \citep{lecun1990handwritten} can be viewed as an {approximation} of PMP \citep{li2017maximum}
opens a room for new optimization algorithms inspired from the optimal control perspective.

\subsubsection{Mean-Field Optimal Control Derivation}
\label{sec:mf-ocp}

In \cite{hu2017control}, a concrete connection was derived between first-order optimization algorithms
and PID controllers.
To see that, consider the formula of gradient descent and the discrete realization of integral control:
\begin{align}
\theta_{k+1} &= \theta_k - \nabla f(\theta_k)                       \label{eq:gd} \\
u_{k+1}      &= \sum\nolimits_{i=0}^{k}     e_i \cdot \Delta t      \label{eq:ic}
\end{align}
These two update rules are equivalent when we interpret the gradient $-\nabla f(\theta_k)$ as tracking error $e(k)$.
Both modules are designed to drive certain statistics of a system,
either the loss gradient or tracking error, towards zero,
by iteratively updating new variables to affect system dynamics.
When a momentum term is introduced, it will result in an additional lag component
which helps increase the low-frequency loop gain \citep{hu2017control}.
This suggests accelerated convergence near local minima,
as observed in the previous analysis \citep{qian1999momentum}.
In other words, the parameters in DNNs can be recast as the control variables in dynamical systems.

With this viewpoint in mind, we can draw
an interesting connection between deep learning training processes and optimal control problems (\ref{eq:ocp}).
In a vanilla discrete-time OCP, the minimization problem takes the form
\begin{equation}
\begin{split}
\min_{%
    \{\vtheta_t\}_{t=0}^{T-1}
}
J :=  \left[
            \Phi(\vx_{T}) + \sum_{t=0}^{T-1}L(\vx_{t}, \vtheta_{t})
     \right] \\
\text{s.t. } \vx_{t+1} = f(\vx_{t}, \vtheta_{t}) \comma
\end{split}
\label{eq:ocp}
\end{equation}
where $\vx_t \in \Xspace $ and $\vtheta_t \in \cspace $ represent state and control vectors.
$f$, $L$ and $\Phi$ respectively denote the dynamics, intermediate cost and terminal cost functions.
In this vein, the goal of (supervised) learning is to find a set of optimal parameters at each time step (i.e. layer),
$\{\vtheta_t\}_{t=0}^{T-1} $,
such that when starting from the initial state $\vx_0$, its terminal state $\vx_T$ is close to the target $\vy$.
Dynamical constraints in \eq{\ref{eq:ocp}} are characterized by the DNNs, whereas
terminal and control-dependent intermediate costs correspond to training loss and weight regularization.
Though state-dependent intermediate costs are not commonly-seen in supervised problems until recently \citep{nokland2019training},
it has been used extensively in the context of deep reinforcement learning
to guide or stabilize training,
e.g. the auxiliary tasks and losses \citep{jaderberg2016reinforcement,liu2017learning}.

Extending \eq{\ref{eq:ocp}} to accept batch data requires viewing the input-output pair $(\vx_0, \vy)$ as a random variable drawn from a probability measure.
This can be done by introducing the mean-field formalism where the analysis is lifted to distribution spaces.
The problem becomes searching an optimal transform that propagates the input population to the desired target distribution.
The population risk minimization problem can hence be regarded as a mean-field optimal control problem \citep{han2018mean},
\begin{equation}
\begin{split}
\inf _{\vtheta_t \in L^{\infty}}
\mathbb{E}_{(\rvx_0, \rvy) \sim \mu_{0}}
    &\left[
        \Phi\left(\rvx_{T}, \rvy \right)
        +\int_{0}^{T} L\left(\rvx_{t}, \vtheta_{t}\right) d t
    \right] \\
\text{s.t. } \dot{\rvx}_{t} =& f(\rvx_{t}, \vtheta_{t}) \comma
\end{split}
\label{eq:mf-ocp}
\end{equation}
where $L^{\infty} \equiv L^{\infty}([0, T], \cspace)$ denotes the set of essentially-bounded measurable functions
and $\mu_0$ is the joint distribution of the initial states $\rvx_0$ and terminal target $\rvy$.
Note that we change our formulation from the discrete-time realization
to the continuous-time framework
since it is mathematically easier to analyze and offers more flexibilities.
The formulation \eq{\ref{eq:mf-ocp}} allows us to analyze the optimization of DNN training through two perspectives,
namely the minimum principle approach and dynamic programming approach, as we will now proceed.

\subsubsection{Mean-Field Pontryagin's Minimum Principle} \label{sec:mf-pmp}
The necessary conditions of the problem \eq{\ref{eq:ocp}} are described in the celebrated
Pontryagin's Minimum Principle (PMP) \citep{boltyanskii1960theory}.
It characterizes the conditions that an optimal state-control trajectory must obey {locally}.
E \textit{et al.} \cite{han2018mean} derived the mean-field extension of the theorem,
which we will restate below.
We will focus on its relation with standard DNN optimization, i.e. gradient descent with back-propagation,
and refer to \cite{han2018mean} for the concrete proof.

\begin{theorem}[Mean-Field PMP \citep{han2018mean}] \label{the:mf-pmp}
Assume the following statements are true:
\begin{itemize} [leftmargin = 32 pt]
    \item [(A1)] The function $f$ is bounded;
                $f$, $L$ are continuous in $\vtheta_t$;
                and $f$, $L$, $\Phi$ are continuously differentiable with respect to $\rvx_t$.
    \item [(A2)] The distribution $\mu_{0}$ has bounded support, i.e.
$\mu_0\left(\left\{(\vx, \vy) \in \Xspace \times \yspace :\|\vx\|+\|\vy\| \leq M\right\}\right)=1$ for some $M>0$.
\end{itemize}
Let $\vtheta_t^*: t \mapsto \cspace $ be a solution that achieves the infimum of \eq{\ref{eq:mf-ocp}}.
Then, there exists continuous stochastic processes $\rvx^*_t$ and $\rvp^*_t$, such that
\begin{alignat}{3}
&\dot{\rvx}_{t}^{*}= \nabla_{\rvp} H\left(\rvx_{t}^{*}, \rvp_{t}^{*}, \vtheta_{t}^{*}\right),
\quad &&\rvx_{0}^{*}=\rvx_{0}
\comma \label{eq:mf-pmp-forward} \\
  &\dot{\rvp}_{t}^{*}=-\nabla_{\rvx} H\left(\rvx_{t}^{*}, \rvp_{t}^{*}, \vtheta_{t}^{*}\right),
\quad &&\rvp_{T}^{*}= \nabla_{\rvx} \Phi\left(\rvx_{T}^{*}, \rvy \right)
\comma \label{eq:mf-pmp-backward}
\end{alignat}
\begin{equation}
\begin{split}
\forall \vtheta_t \in \cspace, &\quad \text{a. e. } t \in[0, T] \comma \\
&\mathbb{E}_{\mu_{0}} H\left(\rvx_{t}^{*}, \rvp_{t}^{*}, \vtheta_t^{*}\right)
    \leq
        \mathbb{E}_{\mu_{0}} H\left(\rvx_{t}^{*}, \rvp_{t}^{*}, \vtheta_t \right)
\comma \label{eq:mf-pmp-max-h}
\end{split}
\end{equation}
where the Hamiltonian function
$H : \Xspace \times \Xspace \times \cspace \rightarrow \mathbb{R} $ is given by
\begin{align}
H\left(\vx_t, \vp_t, \vtheta_t \right) = \vp_t \cdot f(\vx_t, \vtheta_t) + L(\vx_t, \vtheta_t)
\end{align}
and $\vp_t$ denotes the co-state of the adjoint equation.
\end{theorem}

Theorem \ref{the:mf-pmp} resembles the classical PMP result except that
the Hamiltonian minimization condition (\ref{eq:mf-pmp-max-h}) is now taken over an expectation w.r.t. $\mu_0$.
Also, notice the optimal control trajectory $\vtheta^*_t$ admits an open-loop process in the sense that
it does not depend on the population distribution.
This is in contrast to what we will see from the dynamic programming approach (i.e. Theorem \ref{the:mf-hjb}).

The conditions characterized by (\ref{eq:mf-pmp-forward}-\ref{eq:mf-pmp-max-h}) can be linked to the optimization dynamics in DNN training.
First, \eq{\ref{eq:mf-pmp-forward}} is simply the feed-forward pass from the first layer to the last one.
The co-state can be interpreted as the Lagrange multiplier of the objective function w.r.t. the constraint variables \citep{bertsekas1995dynamic}, and its backward dynamics is described in \eq{\ref{eq:mf-pmp-backward}}.
Here, we shall regard \eq{\ref{eq:mf-pmp-backward}} as the back-propagation \citep{li2017maximum}.
To see that, consider the discrete-time Hamiltonian function,
\begin{align}
H \left(\vx_{t}, \vp_{t+1}, \vtheta_{t} \right) = \vp_{t+1} \cdot f(\vx_{t}, \vtheta_{t}) + L(\vx_{t}, \vtheta_{t})\period
\end{align}
Substituting it into the discrete-time version of \eq{\ref{eq:mf-pmp-backward}} will lead to the chain rule used derive back-propagation,
\begin{equation}
\begin{split}
\vp_{t}^{*} &= \nabla_\vx H \left(\vx_{t}^{*}, \vp_{t+1}^{*}, \vtheta_{t}^{*} \right) \\
        &= \vp_{t+1}^{*} \cdot \nabla_\vx f(\vx_{t}^{*}, \vtheta_{t}^{*}) + \nabla_\vx L(\vx_{t}^{*}, \vtheta_{t}^{*}) \comma
\end{split} \label{eq:pmp-back-prop}
\end{equation}
where $\vp_{t}^{*}$ is the gradient of the total loss function w.r.t. the activation at layer $t$.

Finally, the maximization in \eq{\ref{eq:mf-pmp-max-h}} can be difficult to solve exactly
since the dimension of the parameter is typically millions for DNNs.
We can, however, apply approximated updates iteratively using first-order derivatives, i.e.
\begin{align} \label{eq:pmp-gd}
\vtheta^{(i+1)}_t = \vtheta^{(i)}_t - \eta \nabla_{\vtheta_t} \mathbb{E}_{\mu_{0}} H (\vx_{t}^{*}, \vp_{t+1}^{*}, \vtheta^{(i)}_t )\period
\end{align}
The subscript $t$ denotes the time step in the DNN dynamics, i.e. the index of the layer,
whereas the superscript $(i)$ represents the iterative update of the parameters in the outer loop.
The update rule in \eq{\ref{eq:pmp-gd}} is equivalent to performing gradient descent on the original objective function $J$ in \eq{\ref{eq:ocp}},
\begin{align*}
    \nabla_{\vtheta_t} H \left(\vx_{t}^{*}, \vp_{t+1}^{*}, \vtheta_{t} \right)
    &= \vp_{t+1}^{*} \cdot \nabla_{\vtheta_t} f(\vx_{t}^{*}, \vtheta_{t}) + \nabla_{\vtheta_t} L(\vx_{t}^{*}, \vtheta_{t}) \\
    &= \nabla_{\vtheta_t} J \period
\numberthis \label{eq:pmp-gd-explain}
\end{align*}
When the expectation in \eq{\ref{eq:pmp-gd}} is replaced with a sample mean,
E \textit{et al.} \cite{han2018mean} showed that if a solution of \eq{\ref{eq:mf-ocp}} is stable\footnote{
    The mapping $F : U \mapsto V$ is said to be stable on $S_{\rho}(x) :=\left\{y \in U :\|x-y\|_{U} \leq \rho\right\}$
    if $\|y-z\|_{U} \leq K_{\rho}\|F(y)-F(z)\|_{V}$ for some $K_{\rho} > 0$.
},
we can find with high probability a random variable in its neighborhood that is a stationary solution of the sampled PMP.

\subsubsection{Mean-Field Hamilton-Jacobi-Bellman Equation} \label{sec:mf-hjb}
The Hamilton-Jacobi-Bellman (HJB) equation \citep{bellman1964selected} characterizes both necessary and sufficient conditions to the problem \eq{\ref{eq:ocp}}.
The equation is derived from
the principle of Dynamic Programming (DP),
which reduces the problem of minimizing over a sequence of control
to a sequence of minimization over a single control at each time.
This is done by recursively solving the value function (define precisely below), and
the obtained optimal policy is a function applied globally to the state space, i.e. a feedback controller with states as input.

E \textit{et al.} \cite{han2018mean} adapted the analysis to mean-field extension by considering probability measures as states.
Following their derivations,
we will consider the class of probability measures that is square integrable on Euclidean space with $2$-Wasserstein metrics,
denoted $\mathcal{P}_{2}\left(\mathbb{R}^{}\right)$, throughout our analysis.
Importantly, this will lead to an infinite-dimensional HJB equation as we will show later.
It is useful to begin with defining the cost-to-go and value function, denoted as $J$ and $v^*$:
\begin{align}
J(t, \mu, \vtheta_t) :=& %
    \expdnn %
        \left[
            \Phi\left(\rvx_{T}, \rvy \right)
            + \int_{t}^{T} L\left(\rvx_{t}, \vtheta_{t}\right) d t
        \right]
        \label{eq:cost-to-go} \\
v^{*}(t, \mu) :=&
    \inf _{\vtheta_t \in L^{\infty}}
        J(t, \mu, \vtheta_t)
            \label{eq:value}
\end{align}
Note that the expectation is taken over the distribution evolution, starting from $\mu$ and propagating through the DNN architecture.
The cost-to-go function is simply a generalization of the objective \eq{\ref{eq:mf-ocp}} to varying start time,
and its infimum over the control space is achieved by the value function,
i.e. the objective in \eq{\ref{eq:mf-ocp}} can be regarded as $J(0,\mu_0, \vtheta_0)$, with $v^*(0, \mu_0)$ as its infimum.
Now, we are ready to introduce the mean-field DP and HJB equation.
\begin{theorem}[Mean-Field DP \& HJB \citep{han2018mean}] \label{the:mf-hjb}
Let the following statements be true:
\begin{itemize}[leftmargin = 32 pt]
    \item [(A1')] $f$, $L$, $\Phi$ are bounded; $f$, $L$, $\Phi$ are Lipschitz continuous with respect to $\rvx_t$, and the Lipschitz constants of $f$ and $L$ are independent of $\vtheta_t$.
    \item [(A2')] $\mu_{0} \in \mathcal{P}_{2}\left(\xyspace \right)$.
\end{itemize}
Then, both (\ref{eq:cost-to-go}) $J(t, \mu, \vtheta_t)$ and (\ref{eq:value}) $v^{*}(t, \mu)$
are Lipschitz continuous on $[0, T] \times \mathcal{P}_{2}\left(\xyspace \right)$.
For all $0 \leq t \leq \hat{t} \leq T$, the principle of dynamic programming suggests
\begin{align}
v^{*}(t, \mu)=\inf_{\vtheta_t \in L^{\infty}} %
    \expdnn  %
        \left[
            \int_{t}^{\hat{t}} L\left(\rvx_{t}, \vtheta_{t}\right) d t
            + v^{*}\left(\hat{t}, \hat{\mu} \right)
        \right] \comma
 \label{eq:dp}
\end{align}
where $\hat{\mu}$ denotes the terminal distribution at $\hat{t}$.
Furthermore, Taylor expansion of the (\ref{eq:dp}) gives the (\ref{eq:mf-hjb}) equation,
\begin{align}
\left\{\begin{array}{l}{
    {\partial_t v}+\inf _{\vtheta_t \in L^{\infty}}
        \langle \partial_{\mu} v(\mu)(\cdot) , {f}(\cdot, \vtheta_t) \rangle_{\mu}
      + \langle L(\cdot, \vtheta_t)                      \rangle_{\mu}
        = 0,} \\
    {v(T, \mu)=\langle{\Phi}(\cdot) \rangle_{\mu} ,} \end{array}\right.
\label{eq:mf-hjb}
\end{align}
where we recall
$\langle f(\cdot) , g(\cdot) \rangle_{\mu} := \int f(w)^\transpose g(w)\mathrm{d} \mu( w)$
and accordingly denote
$\langle f(\cdot) \rangle_{\mu} := \int f(w) \mathrm{d} \mu( w)$.
Finally, if $\vtheta^*: (t,\mu) \mapsto \cspace$ is a feedback policy
that achieves the infimum in \eq{\ref{eq:mf-hjb}} equation,
then $\vtheta^*$ is an optimal solution of the problem \eq{\ref{eq:mf-ocp}}.
\end{theorem}

Notice that \textit{(A1')} and \textit{(A2')} are much stronger assumptions as opposed to those from Theorem \ref{the:mf-pmp}.
This is reasonable since the analysis is now adapted to take probability distributions as inputs.
Theorem \ref{the:mf-hjb} differs from the classical HJB in that the equations become infinite-dimensional.
The computation requires the derivative of the value function w.r.t. a probability measure,
which can be done by recalling the definition of
the \textit{first-order variation} \citep{ambrosio2008gradient} of a function $F$ at the probability measure $\mu$,
i.e.
$\frac{\partial F (\mu)}{\partial \mu} \equiv \partial_\mu F$,
satisfying the following relation:
\begin{align}
F(\mu + \epsilon f ) = F(\mu) + \epsilon \left\langle \partial_\mu F(\mu)(\cdot) , f(\cdot) \right\rangle_{\mu} + \mathcal{O}(\epsilon^2) \comma
\label{eq:derivative-wrt-density}
\end{align}
where $\epsilon$ is taken to be infinitesimally small.
Note that $F(\cdot)$ and $\partial_\mu F(\mu) (\cdot)$ are functions respectively defined on
the probability measure and its associated sample space.
In other words,
the derivative w.r.t. $\mu$ is achieved by interchanging probability measures with laws of random variables,
to which we can apply a suitable definition of the derivative.

Due to the curse of dimensionality, classical HJB equations can be computationally intractable to solve for high-dimensional problems, let alone its mean-field extension.
However, we argue that in the literature of DNN optimization, algorithms with a DP flavor, or at least an approximation of it,
have not been well-explored.
Research in this direction may provide a principled way to design feedback policies rather than the current open-loop solutions in which weights are fixed once the training ends.
This can be particularly beneficial for problems related to e.g. adversarial attack and generalization analyses.

\section {Stochastic Optimization as a Dynamical System}
\label{sec:sgd}

We now turn attention to stochastic optimization.
Recall that in Sec. \ref{sec:dnn-oc}, we bridge optimization algorithms with controllers.
In classical control theory, controllers alone can be characterized as separated dynamical systems.
Stability analysis is conducted on the compositional system along with the plant dynamics
\citep{franklin1994feedback}.
Similarly, the dynamical perspective can be useful to study the training evolution and convergence property of DNN optimization.
Unlike the deterministic propagation in Sec. \ref{sec:info-prop-dnn},
stochasticity plays a central role in the resulting system due to the mini-batch sampling at each iteration.
Preceding of time corresponds to the propagation of training cycles instead of forwarding through DNN layers.
The stochastic dynamical viewpoint forms most of the recent study on SGD \citep{chaudhari2018stochastic,chaudhari2018deep,an2018stochastic,rotskoff2019neural}.

This section is organized as follows.
In Sec. \ref{sec:pre-mini-batch-grad}, we will review the statistical property of the mini-batch gradient,
which is the foundation for deriving the SGD dynamics.
Recast of SGD as a continuous-time stochastic differential equation (SDE),
or more generally a discrete-time master equation, will be demonstrated in Sec. \ref{sec:conti-time-sgd}, and \ref{sec:discre-time-sgd}, respectively.
The theoretical analysis from the dynamical framework consolidates several empirical observations,
including implicit regularization on the loss landscape \citep{zhang2016understanding}
and phase transition
from a fast memorization period
to slow generalization \citep{shwartz2017opening}.
It can also be leveraged to design optimal adaptive strategies, as we will show in Sec. \ref{sec:sgd-oc-algorithm}.

\subsection {Preliminaries on Stochastic Mini-Batch Gradient}
\label{sec:pre-mini-batch-grad}
Slightly abuse the notation
and denote the averaging training loss on the data set $\mathcal{D}$ as a function of parameter :
\begin{align}
	{\Phi}(\vtheta ; \mathcal{D}) \equiv \Phi(\vtheta) := \frac{1}{\absD} \sum_{i \in \mathcal{D}} J ( f( \rvx^{(i)}, \vtheta ), \rvy^{(i)} ) \comma
	\label{eq:training-loss}
\end{align}
where $J$ is the training objective for each sample (c.f. \eq{\ref{eq:ocp}})
and $f \equiv f_0 \circ f_1 \circ f_2 \cdots$ includes all compositional functions of a DNN.
We can write the full gradient of the training loss as
$\nabla \Phi(\vtheta) \equiv \gradfull = \frac{1}{\absD} \sum_{i \in \mathcal{D}} g^{i}(\vtheta)$,
where $\gradsample$ is the gradient on each data point $(\rvx^{(i)}, \rvy^{(i)})$.
The covariance matrix of $\gradsample$, denoted $\diffusionmatrix$, is a positive-definite (P.D.) matrix
which can be computed deterministically, given the DNN architecture, dataset, and current parameter, as
\begin{align*}
    \Var \left[ g^{i}(\vtheta)\right]
    &:= \tfrac{1}{\absD} \sum\nolimits_{i \in \mathcal{D}}
								 \left(\gradsample - \gradfull\right)\left(\gradsample - \gradfull\right)^{\transpose}
\\ &\equiv \diffusionmatrix \period \numberthis
\end{align*}
Note that in practice, the eigen-spectrum of $\diffusionmatrixx$ often features an extremely low rank
($< 0.5 \% $ for both CIFAR-10 and CIFAR-100 as reported in \cite{chaudhari2018stochastic}).

Access of $\gradfull$ at each iteration is computationally prohibit for large-scale problems.
Instead, gradients can only be estimated through a mini batch of i.i.d. samples $\mathcal{B} \subset \mathcal{D}$.
When the batch size is large enough, $\absB \gg 1$, CLT implies the mini-batch gradient, denoted
$\gradmb = \frac{1}{\absB} \sum_{i \in \mathcal{B}} g^{i}(\vtheta)$,
has the sample mean and covariance
\begin{align*}
	\operatorname{mean}\left[\gradmb\right] :=& \mathbb{E}_{\mathcal{B}}[\gradmb] \approx \gradfull \label{eq:sgd-sample-grad} \numberthis \\ %
	\Var \left[\gradmb\right]  :=& \mathbb{E}_{\mathcal{B}}[\left( \gradmb - \gradfull \right) \left( \gradmb - \gradfull \right)^{\transpose}] \\
											 \approx& \tfrac{1}{\absB} \diffusionmatrix \numberthis \label{eq:sgd-sample-variance} \text{ .}
\end{align*}
The last equality in \eq{\ref{eq:sgd-sample-variance}}
holds when $\mathcal{B}$ is sampled i.i.d. with replacement from $\mathcal{D}$.\footnote{
	$\Var \left[\gradmb\right] = \Var \left[\frac{1}{\absB} \sum_{\mathcal{B}} g^{i}(\vtheta)\right]
								 			= \frac{1}{\absB^2} \sum_{\mathcal{B}} \Var \left[ g^{i}(\vtheta)\right]
								 			= \frac{1}{\absB} \diffusionmatrix$.
	The second equality holds since $\Cov(g^{i}, g^{j}) = 0$ for $i \neq j$.
}
For later purposes, let us also define the two-point noise matrix as $\diffusionmatrixbatch := \mathbb{E}_{\mathcal{B}}[ \gradmb \gradmb^\transpose]$. Its entry $(i,j)$, or more generally the entry $(i_1,i_2, \cdots i_k)$ of a higher-order noise tensor, can be written as
\begin{align}
	\tilde{{\Sigma}}_{\mathcal{B}, (i,j)}                 :=& \mathbb{E}_{\mathcal{B}}[g^{mb}(\theta_{i}) g^{mb}(\theta_j)] \text{ and} \label{eq:ijSimgaB} \\
    \tilde{{\Sigma}}_{\mathcal{B}, (i_1,i_2, \cdots i_k)} :=& \mathbb{E}_{\mathcal{B}}[g^{mb}(\theta_{i_1}) g^{mb}(\theta_{i_2}) \cdots g^{mb}(\theta_{i_k})] \text{ ,} \label{eq:ijkSimgaB}
\end{align}
where $g^{mb}(\theta_i)$ is the partial derivative w.r.t. $\theta_i$ on the mini batch.
Consequently, we can rewrite \eq{\ref{eq:sgd-sample-variance}} as
$\diffusionmatrixbatch - \gradfull \gradfull^\transpose$.

Finally, we will denote the distribution of the parameter at training cycle $t$ as $\rho_t(\vz) := \rho(\vz, t) \propto \mathbb{P}(\vtheta_t=\vz)$ and the steady-state distribution as $\pss := \lim_{t \to \infty } \rho_t$.

\subsection {Continuous-time Dynamics of SGD}
\label{sec:conti-time-sgd}

\subsubsection{Derivation}
The approximations from   (\ref{eq:sgd-sample-grad})-(\ref{eq:sgd-sample-variance}) allow us to replace $\gradmb$ with a Gaussian
$\mathcal{N}\left(\gradfull, \tfrac{1}{\absB} \diffusionmatrixx \right)$.
The updated rule of SGD at each iteration can hence be recast as
\begin{equation}
\begin{split}
\vtheta_{t+1} &=        \vtheta_{t} - \eta \gradmbt \\
              &\approx \vtheta_{t} - \eta \gradfullt + \tfrac{\eta }{\sqrt{\absB}} \diffusionmatrixthaft \mathbf{Z}_t
               \text{ , } \label{eq:approx-sgd}
\end{split}
\end{equation}
where $\eta$ is the learning rate and $\mathbf{Z}_t \sim \mathcal{N}\left( \mathbf{0}, \mathbf{I} \right)$.
Now, consider the following continuous-time SDE and its Euler discretization,
\begin{align}
			 \mathrm{d}\vtheta_{t}   &=                \drift \dt      +                  \diffusion \dWt \\
\Rightarrow            \vtheta_{t+1} &=  \vtheta_{t} + \drift \Delta t + \sqrt{\Delta t} \cdot \diffusion \mathbf{Z}_t \text{ ,} \label{eq:discrete-sde}
\end{align}
In the standard SDE analysis \citep{oksendal2003stochastic}, $\drift$ and $\diffusion$ refer to the drift and diffusion function.
$\dWt$ is the Wiener process, or Brownian motion, in the same dimension of $\vtheta \in \Cspace$.
It is easy to verify that \eq{\ref{eq:discrete-sde}} resembles \eq{\ref{eq:approx-sgd}} if we set
$\Delta t \sim \eta $,
$\drift \sim -\gradfullt $, and
$\diffusion \sim \sqrt{{\eta}/{\absB}} \diffusionmatrixthaftt $.
We have therefore derived the continuous-time limit of SGD as the following SDE,
\begin{align}
			 \mathrm{d}\vtheta_{t}   &= -\gradfullt \dt + \sqrt{2\beta^{-1} \diffusionmatrixt} \dWt \text{ ,}
			 \label{eq:conti-time-sgd}
\end{align}
where $\beta = \frac{2\absB}{\eta}$ is proportional to the inverse temperature in thermodynamics.

Simply viewing \eq{\ref{eq:conti-time-sgd}} already gives us several insights.
First, $\beta$ controls the magnitude of the diffusion process.
A similar relationship, named \textit{noise scale}, between the batch size and learning rate has been proposed in \cite{smith2017bayesian,jastrzkebski2017three}.
These two hyper-parameters, however, are not completely interchangeable since
they contribute to different properties of the loss landscape \citep{wu2018sgd}.
Secondly, the stochastic dynamics is characterized by the drift term from the gradient descent flow $-\gradfullt$ and the diffusion process from $\diffusionmatrixt$.
When the parameter is still far from equilibrium, we can expect the drifting to dominate the propagation.
As we approach flat local minima, fluctuations from the diffusion become significant.
This drift-to-fluctuation transition has been observed on the Information Plane \citep{shwartz2017opening}
and can be derived exactly for convex cases \citep{moulines2011non}.

Since the two Wiener processes in \eq{\ref{eq:conti-time-sgd}} and (\ref{eq:approx-sgd}) are independent,
the approximation is valid only up to the distribution level.
While a more accurate approximation is possible by introducing stochastic modified equations \citep{milstein1994numerical},
we will limit the analysis to \eq{\ref{eq:conti-time-sgd}} and study the resulting dynamics using stochastic calculus and statistical physics.

\subsubsection{Dynamics of Training Loss}
To describe the propagation of the training loss, $\Phi(\vtheta_t)$,
as a function of the stochastic process in \eq{\ref{eq:conti-time-sgd}},
we need to utilize It\^{o} lemma, %
an extension of the chain rule in the ordinary calculus to the stochastic setting:

\begin{lemma}[It\^{o} lemma \citep{ito1951stochastic}] \label{the:ito-lemma}
Consider the stochastic process
$\mathrm{d} X_{t} = {b}\itoarg \dt + {\sigma}\itoarg \mathrm{d} W_{t}$.
Suppose $b(\cdot, \cdot)$ and $\sigma(\cdot, \cdot)$
follow appropriate smooth and growth conditions,
then for a given function $V(\cdot, \cdot) \in C^{2,1}(\mathbb{R}^{d} \times[0, T])$, $V(X_t, t)$ is also a stochastic process:
\begin{align*}
\mathrm{d} V \itoarg
=&\left[ \partial_t V \itoarg + \nabla V \itoarg^{\transpose} b \itoarg \right] \dt \\
 &+ \left[	\frac{1}{2} \Tr \left[ \sigma \itoarg^{\transpose} H_V \itoarg \sigma \itoarg \right] \right] \dt \\
 &+ \left[
	\nabla V \itoarg^{\transpose} \sigma \itoarg
\right] \mathrm{d} W_t
\numberthis
\label{eq:ito} \comma
\end{align*}
where $H_V \itoarg$ denotes the Hessian. i.e.
$H_{V, (i, j)} = \partial^{2} V / \partial x_{i} \partial x_{j}$.
\end{lemma}

Applying \eq{\ref{eq:ito}} to $V=\Phi(\vtheta_t)$
readily yields the following SDE:
\begin{equation}
\begin{split}
\mathrm{d} \Phi\left( \tt \right) =&
	\left[
		- \nabla \Phi( \tt )^{\transpose} g( \tt )
			+\frac{1}{2} \Tr \left[ \diffusionmatrixthafttt \bm{H}_{\Phi} \diffusionmatrixthafttt \right]
	\right] \dt \\
	&+ \left[\nabla \Phi(\tt)^{\transpose} \diffusionmatrixthafttt \right] \dWt \text{ ,}
	\label{eq:loss-sde}
\end{split}
\end{equation}
where we denote $\diffusionmatrixthafttt = \sqrt{2\beta^{-1} \diffusionmatrixt}$ to simplify the notation.
Taking the expectation of \eq{\ref{eq:loss-sde}} over the parameter distribution $\rho_t(\vtheta)$ and recalling $\nabla \Phi = g $,
the dynamics of the expected training loss can be described as
\begin{align}
\mathrm{d} \mathbb{E}_{\rho_t} \left[ \Phi( \tt ) \right] =&
	\mathbb{E}_{\rho_t}
	\left[
		- \nabla \Phi^{\transpose} \nabla \Phi + \frac{1}{2} \Tr \left[ \bm{H}_{\Phi} \diffusionmatrixttt \right]
	\right] \dt \comma
	\label{eq:exp-loss-sde}
\end{align}
which is also known as the backward Kolmogorov equation \citep{kolmogoroff1931analytischen},
a partial differential equation (PDE) that describes the dynamics of a conditional expectation
$\mathbb{E}[f(X_t) | X_s = x]$.
Notice that $\dWt$ does not appear in \eq{\ref{eq:exp-loss-sde}} since the expectation of Brownian motion is zero.
The term $\Tr [ \bm{H}_{\Phi} \diffusionmatrixttt ] $
draws a concrete connection between the noise covariance and the loss landscape.
In \cite{zhu2018anisotropic}, this trace quantity was highlighted as a measurement of the escaping efficiency from poor local minima.
We can also derive the dynamics of other observables following similar steps.

When training converges, the left-hand side of \eq{\ref{eq:exp-loss-sde}} is expected to be near zero, i.e.
\begin{align}
\mathbb{E}_{\pss}[\nabla \Phi^{\transpose} \nabla \Phi] \approx \frac{1}{2} \mathbb{E}_{\pss }[ \Tr [ \bm{H}_{\Phi} \diffusionmatrixttt]] \period
\label{eq:ct-convergence}
\end{align}
In other words, the expected magnitude of the gradient signal is balanced off by the expected Hessian-covariance product measured in trace norm. %
A similar relation can be founded in the discrete-time setting (c.f. \eq{\ref{eq:fdr}}), as we will see later in Sec. \ref{sec:discre-time-sgd}.

\subsubsection{Dynamics of Parameter Distribution}
The dynamics in \eq{\ref{eq:conti-time-sgd}} is a variant of the \textit{damped Langevin diffusion},
which is widely used in statistical physics to model systems interacting with drag and random forces,
e.g. the dynamics of pollen grains suspended in liquid, subjected to the friction from Navier-Stokes equations and random collisions from molecules.
We synthesize the classical results for Langevin systems in the theorem below and discuss its implication in our settings.

\begin{theorem}[Fokker-Plank equation and variational principle \citep{jordan1998variational}] \label{the:langevin}
Consider a damped Langevin dynamics with isotropic diffusion:
\begin{align}
\mathrm{d} X_{t} = -\nabla \Psi(X_t) \dt + \sqrt{2\beta^{-1}} \mathrm{d} W_{t} \comma %
\end{align}
where $\beta$ is the damping coefficient.
The temporal evolution of the probability density,
$\rho_{t} \in C^{2,1}(\mathbb{R}^d \times \mathbb{R}^{+})$,
is the solution of the Fokker-Plank equation (\eq{\ref{eq:fpe}}):
\begin{align}
{\partial_{t} \rho_{t}}
=\nabla \cdot\left(\rho_{t} \nabla \Psi\right)+ \beta^{-1} \Delta \rho_{t}
\label{eq:fpe} \comma
\end{align}
where $\nabla \cdot $, $\nabla$ and $\Delta$ respectively denote the divergence, gradient, and Laplacian operators.
Suppose $\Psi$ is a potential function satisfying appropriate growth conditions,
\eq{\ref{eq:fpe}} has an unique stationary solution given by the Gibbs distribution
$\rho^{\mathrm{ss}}(x; \beta) \propto \exp (-\beta \Psi(x))$.
Furthermore, the stationary Gibbs distribution satisfies the variational principle --- it minimizes the following functional
\begin{align}
\rho^{\mathrm{ss}} = \argmin_{\rho} \mathcal{F}_\Psi(\rho; \beta)
			                      := \mathcal{E}_{\Psi}(\rho) - \beta^{-1}\mathcal{S}(\rho) \comma
			                      \label{eq:vi}
\end{align}
where
$\mathcal{E}_{\Psi}(\rho) := \int_\mathcal{X} \Psi(x) \rho(x) \mathrm{d}x$
and
$\mathcal{S}(\rho) := -\int_{\mathcal{X}} \rho(x) \log \rho(x) \mathrm{d} x$.
In fact,
$\mathcal{F}_\Psi(\rho_t; \beta)$ serves as a Lyapunov function for the \eq{\ref{eq:fpe}},
as it decreases monotonically along the dynamics of FPE and converges to its minimum, which is zero, at $\rho^{\mathrm{ss}}$.
In other words, we  can rewrite \eq{\ref{eq:fpe}} as a form of Wasserstein gradient flow (WGF):
\begin{align}
{\partial_{t} \rho_{t}}
= \nabla \cdot \left(\rho_{t} \nabla (\partial_\rho \mathcal{F}_\Psi)  \right)
\label{eq:wgf-fpe} \comma
\end{align}
where $\partial_\rho \mathcal{F}_\Psi$ follows the same definition in \eq{\ref{eq:derivative-wrt-density}},
and is equal to $\log \frac{\rho}{\rho^{\mathrm{ss}}} + 1$.
We provide the derivation between \eq{\ref{eq:wgf-fpe}} and \eq{\ref{eq:fpe}} in Appendix \ref{app:fpe-derivation}.

\end{theorem}

Equation \eq{\ref{eq:fpe}} characterizes the deterministic transition of the density of an infinite ensemble of particles
and is also known as the forward Kolmogorov equation \citep{oksendal2003stochastic}.
The form of the Gibbs distribution at equilibrium reaffirms the importance of the temperature $\beta^{-1}$,
as it determines the sharpness of $\pss$. %
While high temperature can cause under-fitting,
in the asymptotic limit as $\beta^{-1} \to 0$, the steady-state distribution will
degenerate to point masses located at $\argmax \Psi(x)$.
From an information-theoretic viewpoint, $\mathcal{F}_\Psi(\rho; \beta)$ can be interpreted as the free energy,
where $\mathcal{E}_{\Psi}(\rho)$ and $\mathcal{S}(\rho)$ are respectively known as the energy (or evidence) and entropy functionals.
Therefore, minimizing $\mathcal{F}_\Psi$ balances between the likelihood of the observation and the diversity of the distribution.

Theorem \ref{the:langevin} focuses on the isotropic diffusion process.
Generalization to general diffusion is straightforward, and
adapting the notations from our continuous limit of SGD in \eq{\ref{eq:conti-time-sgd}} yields
\begin{align}
{\partial_t \rho_t}
	= \nabla \cdot\left(\nabla \Phi(\vtheta_t) \rho_t+\beta^{-1} \nabla \cdot( \diffusionmatrixt \rho_t)\right) \comma
\label{eq:fp-sgd}
\end{align}
which is the \eq{\ref{eq:fpe}} of the dynamics of the parameter distribution $\rho_t(\vtheta)$.
Notice that when the analysis is lifted to the distribution space, the drift and diffusion are no longer separable as in \eq{\ref{eq:conti-time-sgd}}.

Now, in order to apply the \eq{\ref{eq:vi}}, %
$\Phi(\vtheta)$ needs to be treated as a potential function under the appropriate growth conditions,
which is rarely the case under the setting of deep learning.
Nevertheless, assuming these assumptions hold will lead to an important implication, %
suggesting that the implicit regularization stemmed from SGD can be mathematically quantified as entropy maximization.
Another implication is that in the presence of non-isotropic diffusion during training\footnote{
	The non-isotropy of $\diffusionmatrixx$ is expected since the dimension of the parameter is much larger than the number of data points used for training.
	Empirical supports can be founded in \cite{chaudhari2018stochastic}.
},
the trajectory governed by \eq{\ref{eq:fp-sgd}} has been shown to converge to a different location from the minima of the training loss \citep{chaudhari2018stochastic}.
In short, the variational inference implied from \eq{\ref{eq:conti-time-sgd}} takes the form
\begin{align}
\argmin_\rho
\mathbb{E}_{\vtheta \sim \rho_t}[\tilde{\Phi} (\vtheta)]-\beta^{-1} \mathcal{S}(\rho_t) \comma
\end{align}
which is minimized at $\rho^{\mathrm{ss}}(\vtheta) \propto \exp({-\beta \tilde{\Phi}(\vtheta)})$.
The relationship between ${\Phi}$ and $\tilde{\Phi}$
at equilibrium is given by\footnote{
    The derivation of \eq{\ref{eq:vi-training-loss}} is quite involved as it relies on the equivalence between It\^{o} and A-type stochastic integration for the same FPE. We refer readers to \cite{chaudhari2018stochastic} for a complete treatment.
}
\begin{align}
\nabla \Phi= \diffusionmatrixx \nabla \tilde{\Phi}-\beta^{-1} \nabla \cdot \diffusionmatrixx \label{eq:vi-training-loss} \comma
\end{align}
where the divergence $\nabla \cdot \diffusionmatrixx$ is applied to the column space of the diffusion matrix.
In other words, the critical points of $\tilde{\Phi}$ differ from those of the original training loss by the quantity
$\beta^{-1} \nabla \cdot \diffusionmatrixx$.
It can be readily verified that %
$\tilde{\Phi} = \Phi$ if and only if $\diffusionmatrixx$ is isotropic, i.e.
$\diffusionmatrixx = c \mI_{\Cspace \times \Cspace }$ for some constant $c$.
In fact, we can construct cases in which the most-likely trajectories traverse along closed loops, i.e. limit cycles, in the parameter space \citep{chaudhari2018stochastic}.

\subsubsection{Remarks on Other SDE Modeling}
We should be aware that \eq{\ref{eq:conti-time-sgd}}, as a variant of the well-known Langevin diffusion,
is only one of the
possible realization of modeling stochastic mini-batch gradient. %
In fact, the metastability analysis of Langevin diffusion \citep{bovier2004metastability}
conflicts with empirical observations in deep learning,
as the analysis suggests an escape time depending exponentially on the depth of the loss landscape but only polynomial with its width.
In other words, theoretical study implies Brownian-like processes should stay much longer in sharp local minima.
To build some intuition on why this is true, recall that
an implicit assumption we made when deriving \eq{\ref{eq:conti-time-sgd}} is the finite variance induced by $\gradmb$.
Upper-bounding the second moment eventually prevents the presence of long-tail distributions,
which plays a pivotal role in speeding up
the exponential exit time of an SDE from narrow basins.

This issue has been mitigated in \cite{simsekli2019tail}
by instead considering a general L\^{e}vy process:
\begin{align}
	 \mathrm{d}\vtheta_{t}   &= -\gradfullt \dt + \eta^{\frac{\alpha-1}{\alpha}} \sigma_\alpha(\vtheta_t)  \dLat \text{ ,}
	 \label{eq:conti-time-sgd-levy}
\end{align}
where $\alpha \in (0,2]$ is the tail index and $\dLat$ denotes the $\alpha$-stable L\^{e}vy motion.
The mini-batch gradients are now drawn from a zero-mean symmetric $\alpha$-stable L\^{e}vy distribution,
$\mathcal{S}\alpha\mathcal{S}$-$\mathrm{Levy}(0, \sigma_\alpha)$.
Note that the moment of the distribution $\mathcal{S}\alpha\mathcal{S}$-$\mathrm{Levy}$ is finite up to only $\alpha$ order.
When $\alpha = 2$, $\mathcal{S}\alpha\mathcal{S}$-$\mathrm{Levy}$ degenerates to a Gaussian
and $\dLat$ is equivalent to a scaled Brownian motion.
On the other hand,
for $\alpha < 2$, the stochastic process in \eq{\ref{eq:conti-time-sgd-levy}}
features a Markov ``jump'' behavior,
and theoretical study indicates a longer stay in,
i.e. the process prefers, wider minima valleys \citep{bovier2004metastability}.
The resulting heavy-tailed density aligns better with the empirical observation \citep{chaudhari2018stochastic}.

\subsection {Discrete-time Dynamics of SGD}
\label{sec:discre-time-sgd}
Despite the rich analysis by formulating SGD as a continuous-time SDE,
we should remind us of the implicit assumptions for It\^{o}-Stratonovich calculus to apply.
Beside the smoothness conditions on the stochastic process, mini-batch gradients are replaced with Gaussian to bring Brownian motion into the picture.
The fact that the mean square displacement of Brownian motion scales linearly in time, i.e.
$\mathbb{E}[\dWt^2] = \dt$, leads to a quadratic expansion on the loss function, as shown in \eq{\ref{eq:loss-sde}}.
In addition, the recast between \eq{\ref{eq:approx-sgd}} and (\ref{eq:conti-time-sgd}) requires splitting $\eta$ to $\sqrt{\eta} \sqrt{\dt}$.
The continuous limit reached by sending $\dt \to 0^{+}$ while assuming finite $\sqrt{\eta}$ is arguably unjustified and pointed out in \cite{yaida2018fluctuation}.
The authors instead proposed a discrete-time master equation that alleviates these drawbacks
and is able to capture higher-order structures.
We will restate the result and link it to the continuous-time SDE formulation as proceeding.

\subsubsection{Derivation}
Recall that $\rho_t(\vtheta)$ is the distribution of the parameter at time $t$.
Its dynamics, when propagating to the next cycle $t+1$, can be written generally as
\begin{align}
\rho_{t+1}(\vtheta) = \mathbb{E}_{\thprime \sim \rho_t, \mathcal{B}}	\left[ \deltaa \right] \comma
 \label{eq:discrete-dist-sgd}
\end{align}
where $\delta\{\cdot\}$ is the Kronecker delta and the expectation is taken over both the current distribution and the mini-batch sampling.
Given an observable $\mathcal{O}(\vtheta): \Cspace \mapsto \mathbb{R}$,
its master equation at steady-state equilibrium $\pss$ can be written as
\begin{align}
\expPss \left[ \mathcal{O}(\vtheta) \right]
	= \expPss \left[ \expB \left[ \mathcal{O}\left(\gmbupdatee\right) \right] \right] \label{eq:master} \period
\end{align}
Full derivations are left in Appendix \ref{app:master-eq-derive}.
Note that the only assumption we make so far is the existence of $\pss$.
Equation \eq{\ref{eq:master}} suggests that at equilibrium,
the expectation of an observable remains unchanged when
averaging over the stochasticity from mini-batch gradient updates.

\subsubsection{Fluctuation Dissipation of Training Loss}
Let we proceed by plugging the training loss $\Phi(\vtheta)$ to our observable of interest in \eq{\ref{eq:master}}.
Taylor expanding it at $\eta = 0$ gives
\begin{align*}
&\expPss \left[ \Phi(\vtheta) \right] \\
	=& \expPssof{\expBof{\Phigmd}} \numberthis \label{eq:tayer} \\
	=& \expPssof{\Phi(\vtheta) + \sum_{k=1}^{\infty} \frac{(-\eta)^{k}}{k !} \Dk \expBof{\Phigmd}}
	\comma
\end{align*}
where $\Dk F$ denotes the $k$-ordered expansion on a multivariate function $F$.
Specifically,
the first and second expansions can be written as\footnote{
Applying the chain rule to $\Dk \expBof{\cdots}$ in \eq{\ref{eq:tayer}} yields a clean form as
$
\Dk \expBof{\Phigmd} = \sum \mathrm{D}^k_{(i_1,i_2, \cdots i_k)} \Phi(\vtheta) \cdot
												   \tilde{\mathbf{\Sigma}}_{\mathcal{B}, (i_1,i_2, \cdots i_k)}
$,
where we recall \eq{\ref{eq:ijkSimgaB}} and
denote $\mathrm{D}^k_{(i_1,i_2, \cdots i_k)}$ as the $k$-ordered partial derivatives w.r.t. parameter indices $i_1,i_2, \cdots i_k$.
For instance, $\mathrm{D}^1_{(i)} \Phi(\vtheta) = \partial_{\theta_i} \Phi $ corresponds to the $i$-element of the full gradient.
$\mathrm{D}^2_{(i,j)} \Phi(\vtheta) = \partial^{2} \Phi / \partial \theta_{i} \partial \theta_{j} $
refers to the $(i,j)$-entry of the Hessian.
The summation $\sum$ is taken over combinations of indices $i_1,i_2, \cdots i_k$.
\label{footnote1}
}
\begin{align*}
\mathrm{D}^1 \expBof{\Phigmd}
=& \smallsum_{i} \partial_{\theta_i} \Phi \cdot \expBof{g^{mb}(\theta_i)} \\
=& \nabla \Phi^\transpose \nabla \Phi \numberthis \label{eq:d1} \\
\mathrm{D}^2 \expBof{\Phigmd}
=& \smallsum_{(i, j)} H_{\Phi, (i, j)} \tilde{\mathbf{\Sigma}}_{\mathcal{B}, (i, j)} \\
=& \Tr   \left[ \mH_\Phi \diffusionmatrixbatch \right] \numberthis \period \label{eq:d2}
\end{align*}
For the last equality to hold in \eq{\ref{eq:tayer}}, the expectation of the infinite-series summation needs to vanish.
Substituting   (\ref{eq:d1}-\ref{eq:d2}) to (\ref{eq:tayer}), we will obtain the following relation
\begin{align*}
\expPss \left[\nabla \Phi^{\transpose} \nabla \Phi \right]
	&= \frac{\eta}{2} \expPss \left[ \Tr   \left[ \mH_\Phi \diffusionmatrixbatch \right] \right] \numberthis \label{eq:fdr} \\
	&+ \sum_{k=3}^{\infty} \frac{(-\eta)^{k}}{k !} \mathbb{E}_{\pss, \mathcal{B}} \left[ \Dk \Phi(\gmbupdatee) \right] \period
\end{align*}
\eq{\ref{eq:fdr}} can be viewed as the fluctuation-dissipation equation,
a key concept rooted in statistical mechanics for bridging microscopic fluctuations to macroscopic dissipative phenomena \citep{yaida2018fluctuation}.
It should be noted, however, that \eq{\ref{eq:fdr}} is the necessary but not sufficient condition to ensure stationary.

Let us compare \eq{\ref{eq:fdr}} with its continuous-time counterpart in \eq{\ref{eq:ct-convergence}}.
First, notice the difference between the two-point noise matrix, $\diffusionmatrixbatch$,
and the covariance matrix of the sample gradient, $\diffusionmatrixttt$.
In fact, these two matrices can be related by
\begin{align}
\diffusionmatrixbatch
\approx \left( \tfrac{1}{\absB} - \tfrac{1}{\absD} \right)\diffusionmatrixx
= 		\tfrac{1}{\eta}\left( 1 - \tfrac{\absB}{\absD} \right) \diffusionmatrixttt
\approx \tfrac{1}{\eta}\diffusionmatrixttt \comma
\end{align}
where the first approximation is followed by Proposition 1 in \cite{hu2017diffusion}, and the second one by assuming $\absB \ll \absD$.
In the small learning rate regime, \eq{\ref{eq:ct-convergence}} and (\ref{eq:fdr}) are essentially equivalent
and we consolidate the analysis from the continuous-time framework in Sec. \ref{sec:conti-time-sgd}.
The higher-order terms in \eq{\ref{eq:fdr}} measure the anharmonicity of the loss landscape,
which becomes nontrivial when the learning rate is large.

\subsection {Improving SGD with Optimal Control}
\label{sec:sgd-oc-algorithm}

Interpreting SGD as a stochastic dynamical system makes control theory applicable.
Note that this is different from what we have derived in Sec. \ref{sec:dnn-dynamics}.
The state space on which we wish to impose control is the parameters space $\Cspace$, instead of the activation space $\Xspace$.
The fact that in Sec. \ref{sec:dnn-oc} we are managing to apply control in $\Cspace$
limits out capability to go beyond theoretical characterization to practical algorithmic design due to high dimensionality.
In contrast, here we do not specify where the control should take place,
depending on how we introduce it to the stochastic dynamical system.
Such a flexibility has led to algorithmic improvement of SGD dynamics.
For instance, using optimal control to derive optimal adaptive strategies for hyper-parameters \citep{li2017stochastic}.

The literature on adaptive (e.g. annealed) learning rate scheduling has been well-studied for convex problems \citep{moulines2011non,xu2011towards}.
The heuristic of decaying the learning rate with training cycle, i.e. $\sim 1/t$,
typically works well for DNN training, despite its non-convexity.
From the optimal control perspective, we can formulate the scheduling process mathematically by introducing to \eq{\ref{eq:conti-time-sgd}} a rescaling factor $u_t \in (0,1]$ and its continuous-time function $u_{t\to T} : [t,T] \mapsto (0,1]$.
Applying similar derivations using It\^{o} Lemma, we obtain
\begin{align}
			 \mathrm{d}\vtheta_{t}   &= - u_t \gradfullt \dt + u_t \sqrt{2\beta^{-1} \diffusionmatrixt} \dWt \comma  \label{eq:control-sgd} \\
\mathrm{d} \mathbb{E} \left[ \Phi( \tt ) \right] &=
	\mathbb{E}
	\left[
		- u_t \nabla \Phi^{\transpose} \nabla \Phi + \frac{u_t^2}{2} \Tr \left[ \mH_{\Phi} \diffusionmatrixttt \right]
	\right] \dt \text{.} \label{eq:control-exp-loss-sde}
\end{align}
The \textit{stochastic optimal control problem} from this new dynamics can be written as
\begin{align}
	\min_{u_{t\to T}} \mathbb{E} \left[ \Phi( \vtheta_T ) \right] \quad \text{s.t. (\ref{eq:control-exp-loss-sde})} \comma \label{eq:ocp-sgd}
\end{align}
where $T$ is the maximum training cycle.
Li \textit{et al.} \cite{li2017stochastic} showed that when $\Phi( \vtheta )$ is \textit{quadratic},
solving \eq{\ref{eq:ocp-sgd}} using the HJB equation (recall Theorem \ref{the:mf-hjb}) yields a closed-form policy
\begin{align}
	u_t^{*} = \min (1, \frac{\mathbb{E} [ \Phi( \tt ) ]}{\eta \diffusionmatrixttt}) \period \label{eq:ocp-sgd-convex}
\end{align}
Intuitively, this optimal strategy suggests using the maximum learning rate when far from minima
and decay it whenever fluctuations begin to dominate.
Further expansion on the ratio $\mathbb{E} [ \Phi( \tt ) ] / \eta \diffusionmatrixttt$ will give us the annealing schedule of $\mathcal{O}(1/t)$.
In other words, the strategy proposed in the previous study is indeed optimal from the optimal control viewpoint.
Also, notice that \eq{\ref{eq:ocp-sgd-convex}} is a feedback policy since the optimal adaptation depends on the statistics of the current parameter.

For general loss functions, \eq{\ref{eq:ocp-sgd-convex}} can serve as a well-motivated heuristic.
The resulting scheduling scheme has been shown to be more robust to initial conditions when compared with other SGD variants \citep{li2017stochastic}.
Similarly,
we can derive optimal adaptation strategies for other hyper-parameters, such as the momentum and batch size \citep{li2017stochastic,an2018stochastic}.
Lastly, we note that other learning rate adaptations, such as the constant-and-cut scheme,
can also be included along this line by modifying \eq{\ref{eq:control-sgd}} to accept general Markov jump processes.

\section {Beyond Supervised Learning}
\label{sec:beyond-sl}

The optimal control framework in Sec. \ref{sec:dnn-oc}
fits with supervised learning by absorbing labels into the terminal cost or augmented state space.
In this section, we demonstrate how to extend the framework to other learning problems.
Specifically, by allowing standard (i.e. risk-neutral) \eq{\ref{eq:ocp}} and \eq{\ref{eq:mf-ocp}} objectives,
which minimize the expected loss incurred from the stochasticity,
to be risk-aware,
we generalize the formulation to consider statistical behaviors from higher-order moments.
Depending on the problem setting, the risk-aware optimal control problem can be recast to Bayesian learning and adversarial training,
as we will show in Sec. \ref{sec:bayesian-learning} and \ref{sec:adversarial-learning}.
While the former viewpoint has been leveraged to impose priors on the training dynamics \citep{chaudhari2016entropy,chaudhari2018deep},
the latter seeks to optimize worst-case perturbations from an adversarial attacker.
Additionally, we will interpret meta-learning algorithms with a specific structure as feedback controllers in Sec. \ref{sec:meta-learning}.

\subsection {Preliminaries on Risk-Aware Optimal Control}
\label{sec:pre-risk-aware}
Risk sensitivity has been widely used in Markovian decision processes (MDPs)
that require more sophisticated criteria to reflect the variability-risk features of the problems \citep{coraluppi1999risk}.
The resulting optimal control framework is particularly suitable for stochastic dynamical systems and
closely related to robust and minimax control \citep{bacsar2008h}.
To bring risk awareness into the original training objective, i.e. the per-sample objective $J$ in \eq{\ref{eq:ocp}}, we need to consider the following generalized exponential utility function:
\begin{align}
\mathcal{J}_k(\vx, \xi):=\left\{
	\begin{array}{ll}
		{\frac{1}{k} \log \left\{\mathbb{E}_{\xi}\left[\exp \left(k J(\vx, \xi)\right)\right]\right\}} & {, k \neq 0} \\
		{\mathbb{E}_\xi\left[J(\vx, \xi) \right]}                                & {, k=0}
	\end{array}\right.
	, \label{eq:risk-aware-obj}
\end{align}
where $\xi$ denotes any source of stochasticity that is being averaged over the expectation.
When $k=0$, $\mathcal{J}_k$ reduces to the risk-neutral empirical mean, i.e. $\Phi(\vtheta)$ in \eq{\ref{eq:training-loss}}.
In contrast, the log partition functional for $k \neq 0$ has a risk-aware interpretation, which can be mathematically described as
\begin{align}
\mathcal{J}_{k\neq 0}(\vx, \xi) \approx
\mathbb{E}_\xi J + \frac{k}{2} \Var_\xi \left[ J \right] \period \label{eq:risk-variance-obj}
\end{align}
We left the full derivation in Appendix \ref{app:minimax-logsumexp}.
For positive $k$, the objective is \textit{risk-averse} since in addition to the expectation,
we also penalize the variation of the loss.
In contrast, $k < 0$ results in a \textit{risk-seeking} behavior as we now favor higher variability.
From the optimization viewpoint, the log partition functional can be thought of as an approximation of a smooth $\max$ operator.
The objective in \eq{\ref{eq:risk-aware-obj}} therefore inherits an inner-loop $\max$/$\min$ optimization, depending on the sign of $k$:
\begin{align}
\min_{\vx} \mathcal{J}_{k\neq 0}(\vx, \xi) \approx \left\{
	\begin{array}{lll}
		{\displaystyle\min_{\vx} \max_{\xi} J(\vx, \xi) } & {\text { if  }} & { k > 0} \\
		{\displaystyle\min_{\vx} \min_{\xi} J(\vx, \xi) } & {\text { if  }} & { k < 0}
	\end{array}\right.
	 \label{eq:minmax-minmin}
\end{align}
From such, it is handy to characterize the optimal policy of a $\min$-$\max$ objective as risk-averse, whereas
the one from a $\min$-$\min$ objective often reveals a risk-seeking tendency.
This interpretation will become useful as we proceed to Sec. \ref{sec:bayesian-learning} and \ref{sec:adversarial-learning}.

\subsection{Bayesian Learning \& Risk-Seeking Control}
 \label{sec:bayesian-learning}
Recall that flatter minima enjoy lower generalization gap since they are less sensitive to perturbations of the data distribution \citep{zhang2016understanding}.
In this spirit, Chaudhari \textit{et al.} \cite{chaudhari2016entropy} proposed the following \textit{local entropy loss}
in order to guide the SGD dynamics towards flat plateaus faster:
\begin{align}
\Phi_{\text{ent}}(\vtheta ; \gamma) := \LocalEntropy \comma \label{eq:local-entropy}
\end{align}
where $\Phi (\cdot)$ is defined in \eq{\ref{eq:training-loss}} and
the hyper-parameter $\gamma$ controls the degree of trade-off between the depth and width of the loss landscape.
This surrogate loss is well-motivated from the statistical physics viewpoint since
the objective balances between an energetic term (i.e. training loss) and entropy term (i.e. flatness of local geometry).
In addition, it can be connected to numerical analysis on nonlinear PDE \citep{chaudhari2018deep}.
Here, we provide an alternative perspective from risk-aware control and its connection to Bayesian inference.

We know from Sec. \ref{sec:pre-risk-aware} that the log partition functional approximates the $\max$ operator.
Minimizing the local entropy loss therefore becomes
\begin{align*}
			\min_{\vtheta} \Phi_{\text{ent}}(\vtheta ; \gamma)
\approx&	\min_{\vtheta} - \max_{\thetap} \left\{ - \Phi (\thetap)-\frac{\gamma}{2}\|\vtheta-\thetap\|_{2}^{2} \right\} \\
=&	\min_{\vtheta}   \min_{\thetap} \left\{   \Phi (\thetap)+\frac{\gamma}{2}\|\vtheta-\thetap\|_{2}^{2} \right\} \numberthis \label{eq:sgd-min-min} \comma
\end{align*}
which is a nested optimization
with an inner loop minimizing the same loss with a regularization term centered at $\vtheta$.
For fixed $\thetap$, the outer loop simply optimizes a locally-approximated quadratic
$\frac{\gamma}{2}\|\vtheta-\thetap\|_{2}^{2}$.
Casting this quadratic regularization as a distribution density and recalling the risk-seeking interpretation of the $\min$-$\min$ objective,
we have
\begin{align}
\min_{\vtheta} \Phi_{\text{ent}}(\vtheta ; \gamma)
\approx
\min_{\vtheta}
\mathbb{E}_{\mathcal{P}_{\gamma, \vtheta}} \left[ \Phi (\thetap) \right] - \frac{1}{2} \Var_{ \mathcal{P}_{\gamma, \vtheta}} \left[ \Phi (\thetap) \right] \label{eq:risk-seek-sgd} \period
\end{align}
$\mathcal{P}_{\gamma, \vtheta}$ denotes the Gibbs distribution,
$ \mathcal{P}(\thetap ; \gamma, \vtheta) \propto Z^{-1} \exp (-\frac{\gamma}{2}\|\vtheta-\thetap\|_{2}^{2}) $,
with $Z^{-1}$ as the normalization term.
The risk-seeking objective in \eq{\ref{eq:risk-seek-sgd}} encourages exploration on areas with higher variability.
This implies a potential improvement on the convergence speed, despite the overhead incurred from additional minimization.

Solving \eq{\ref{eq:sgd-min-min}} requires an expensive inner-loop minimization over the entire parameter space at each iteration.
This can, however, be estimated with the stochastic gradient Langevin dynamic \citep{welling2011bayesian}, an MCMC sampling technique for Bayesian inference.
The resulting algorithm, named \textit{Entropy-SGD} \citep{chaudhari2016entropy}, obeys the following dynamics:
\begin{align}
	 \mathrm{d}\vz_{s}  &= - \left[ g^{mb}(\vz_s) + \gamma \left(\vz_s - \vtheta_t \right) \right] \ds + \sqrt{\epsilon} \mathrm{d} \mathbf{W}_s \comma \label{eq:inner-entropy-sgd} \\
	 \mathrm{d}\vtheta_{t}   &= \gamma \left( \vz - \vtheta_t \right) \dt \comma \label{eq:outer-entropy-sgd}
\end{align}
where $\vz$ takes the same space as $ \vtheta \in \Cspace$ with the initial condition $\vz_{0} = \vtheta_t$.
Notice that the two dynamical systems in \eq{\ref{eq:inner-entropy-sgd}} and (\ref{eq:outer-entropy-sgd})
operate in different time scales, denoted $\ds$ and $\dt$ respectively,
and they correspond to the first-order derivatives of the inner and outer minimization in \eq{\ref{eq:sgd-min-min}}.
Chaudhari \textit{et al.} \cite{chaudhari2018deep} showed that in the asymptotic limit of the non-viscous approximation, i.e. $\epsilon \to 0$,
gradient descent on the local entropy loss,
$\vtheta_{t+1} \leftarrow \vtheta_t - \eta \nabla_{\vtheta} \Phi_{\text{ent}}(\vtheta_t ; \gamma) $, is equivalent to
a \textit{forward} Euler step on the original loss function,
$\vtheta_{t+1} \leftarrow \vtheta_t - \eta \nabla_{\vtheta} \Phi(\vtheta_{t+1}) $.
In other words, we may interpret the dynamics in \eq{\ref{eq:inner-entropy-sgd}} as a one-step prediction
(in a Bayesian fashion with a quadratic prior)
of the gradient at the next iteration.

\subsection{Adversarial Training as Minimax Control} \label{sec:adversarial-learning}
Study on adversarial properties of DNN has become increasingly popular %
since the landmark paper in \cite{szegedy2013intriguing} revealed its vulnerability to human-invisible perturbations.
The consequence can be catastrophic from a security standpoint \citep{goodfellow2018defense},
when machine learning algorithms in real-world applications, e.g. perception systems on self-driving vehicles,
are intentionally fooled (i.e. attacked) to make incorrect decisions at test time.
Among the attempts to robustify deep models,
adversarial training
proposes to solve the following optimization problem:
\begin{align}
	\min_{\vtheta} \max_{\lVert\delta\rVert_{p} \leq \Delta} \Phi_{\text{adv}}(\vtheta, \delta)
	:= \frac{1}{\absD} \sum_{i \in \mathcal{D}} J ( f( \rvx^{(i)} + \delta^{(i)}, \vtheta ), \rvy^{(i)} )
	\label{eq:minimax-adv-training} \period
\end{align}
$\Phi_{\text{adv}}(\vtheta, \delta)$ is equivalent to the original training loss
(c.f. \eq{\ref{eq:training-loss}})
subjected to sample-wise perturbations $\delta^{(i)}$,
which are of the same dimension as the input space and
constrained within a $p$-norm ball with radius $\Delta$.
Essentially, adversarial training seeks to find a minimizer of the worse-case performance when data points are adversarially distorted.

The $\min$-$\max$ objective in \eq{\ref{eq:minimax-adv-training}} implies a risk-averse behavior,
in contrast to the risk-seeking in \eq{\ref{eq:risk-seek-sgd}}.
Classical analyses from the minimax control theory suggest a slow convergence and a conservative optimal policy.
These arguments agree with practical observations as adversarial learning usually takes much longer time to train and admits a trade-off between adversarial robustness (e.g. proportion of data points that are adversarial) and generalization performance
\citep{goodfellow2018defense}.

Algorithmically,
the inner maximization is often evaluated on a set of adversarial examples generated on the fly, depending on the current parameter,
and the adversarial training objective is replaced with a mixture of the original loss function and this adversarial surrogate.
The $\min$-$\max$ problem in \eq{\ref{eq:minimax-adv-training}} is hence lower-bounded by
\begin{equation}
\begin{split}
	\min_{\vtheta} \max_{\lVert\delta\rVert_{p} \leq \Delta} \Phi_{\text{adv}}(\vtheta, \delta)
\geq\min_{\vtheta} \alpha \Phi(\vtheta) + (1-\alpha)		 \Phi_{\text{adv}}(\vtheta, \hat{\delta}) \comma
	 \label{eq:adv-training}
\end{split}
\end{equation}
where $\alpha \in (0,1]$ is the mixture ratio and
$\hat{\delta} := \Proj_{\lVert\cdot\rVert_{p} \leq \Delta} [ \Alg(\vtheta, \Phi(\cdot); \mathcal{D})]$
denotes the $p$-norm projected perturbation generated from an algorithm, $\Alg$.
The approach can be viewed as an adaptive
data augmentation technique.
We should note, however, that the i.i.d. assumption on the training dataset no longer holds in this scenario and we may require exponentially more data to prevent over-fitting \citep{schmidt2018adversarially}.
Lastly, it is possible to instead upper-bound the objective with a convex relaxation,
which will lead to a provably robust model to any norm-bounded adversarial attack \citep{wong2017provable}.

\subsection{Meta Learning as Feedback Controller} \label{sec:meta-learning}

Meta-learning aims to discover prior knowledge from a set of learnable tasks
such that the learned initial parameter provides a favorable inductive bias for fast adaptation to unseen tasks at test time.
The learning problems, often called \textit{learning to learn},
is naturally applicable to those involving prior distillation from limited data,
such as few-shot classification \citep{finn2017model} and reinforcement learning in fast-changing environments \citep{duan2016rl}.
It can also be cast to probabilistic inference in a hierarchical Bayesian model \citep{grant2018recasting}.
Here, we bridge a popular branch of algorithms, namely \textit{model-agnostic meta-learning} (MAML) \citep{finn2017model},
to the feedback control viewpoint.

In the problem formulation of MAML, an agent is given a distribution of tasks, $\mathcal{T}_i \sim \mathcal{P}_\mathcal{T}$,
with the task-dependent cost function, $\Phi_{\mathcal{T}_i}(\cdot)$,
and asked to find an initialization that can continuously adapt to other unseen tasks drawn from
$\mathcal{P}_\mathcal{T}$.
The meta-training objective and adaptation rule can be written as
\begin{align}
	\Phi_{\text{meta}}(\vtheta; \mathcal{P}_\mathcal{T})
	:=& \mathbb{E}_{\mathcal{T}_i \sim \mathcal{P}_\mathcal{T}}
		\left[
			\Phi_{\mathcal{T}_i}(\vtheta_{\text{adapt}}^N)
		\right]
	\text{, where } \label{eq:meta-learning-obj} \\
	\metanext = \metacurr - \bar{\eta} \nabla_\vtheta &\Phi_{\mathcal{T}_i}(\metacurr) \quad
	\text{and} \quad    \vtheta_{\text{adapt}}^{0} = \vtheta
	\period \label{eq:meta-adapt-rule}
\end{align}
$\vtheta_{\text{adapt}}^N$ denotes an $N$-step adaptation from the current parameter
using gradient descent with the step size $\bar{\eta}$ at each update.
$N$ is a hyper-parameter
that generalizes standard objectives to $\Phi_{\text{meta}}(\cdot)$ for positive $N$.
As $N$ increases, regularization will be imposed on the meta-training process
in the sense that the agent is encouraged to find a minimizer
no more than $N$ steps away from the local minima of each task,
instead of over-fitting to the one of any particular task.

Now, recall the interpretation of (\ref{eq:gd}) as (\ref{eq:ic}) in Sec. \ref{sec:dnn-oc}.
Through this lens, the adaptation rule in \eq{\ref{eq:meta-adapt-rule}}
can be thought of as an $N$-step integral controller,
and minimizing \eq{\ref{eq:meta-learning-obj}} is equivalent to searching
an optimal initial condition for the controller. %
Since feedback controllers are originally designed for problems requiring on-line adaptation
and integral controllers feature zero steady-state errors,
we consolidate the theoretical foundation of MAML-inspired algorithms.
Implications from this viewpoint can leverage knowledge from control literature
to design more sophisticated and/or principled adaptation rules.
We may also derive optimal adaptation rules for other hyper-parameters,
such as the step size $\bar{\eta}$ and adaptation number $N$,
similar to what we have shown in Sec. \ref{sec:sgd-oc-algorithm}.

\section {Conclusion}
\label{sec:diss-conclusion}

This review aims to align several seemly disconnected viewpoints of deep learning theory with the line of
dynamical system and optimal control.
We first observe that the compositionality of DNNs and the descending update in SGD
suggest an interpretation of discrete-time (stochastic) dynamical systems.
Rich mathematical analysis can be applied
when certain assumptions are made to bring the realization to its continuous-time limit.
The framework
forms the basis of most recent understandings of deep learning,
by recasting DNN as an ordinary differential equation and SGD as a stochastic differential equation.
Among the available {mathematical} tools,
we should highlight the significance of {mean-field theory} and {stochastic calculus}, %
which enable characterization of the dynamics of deep representation and stochastic functionals (e.g. the training loss or parameter distribution) at the ensemble level.
The dynamical perspective alone has revealed valuable implications, as it successfully gives predictions to e.g.
the trainability of random networks from critical initialization,
the interaction between gradient drift and noise diffusion during training,
the concrete form of implicit regularization from SGD,
and even the global optimality of deep learning problems,
to name a few.

Another appealing implication, despite receiving little attention, is to introduce
the optimal control theory to the corresponding dynamics.
To emphasize its importance, we note that
the celebrated back-propagation algorithm
is, in fact, an approximation of the Pontryagin’s Minimum Principle (PMP),
a well-known {theory} dated back to the 1960s that describes the {necessary} conditions to the optimal control problems.
Limited works {inspired from this viewpoint} include optimal adaptive strategies for hyper-parameters
and minimum principle based optimization algorithms.
When the standard optimal control objective is extended to accept higher-order statistical moments,
the resulting ``risk-aware'' optimal control framework generalizes beyond supervised learning, to include problems such as Bayesian learning, adversarial training, and meta learning.
We wish this article stands as a foundation to open up new avenues that may bridge and benefit communities from both deep learning and optimal control.

{
For future directions, we note that
the optimal control theory for DNNs training is far from being completed,
and relaxing the currently presented theorems to a more realistic setting will be beneficial.
For instance,
despite the thorough discussion in Sec. \ref{sec:dnn-oc},
our derivation is mainly {constructed} upon the continuous-time framework
to avoid the difficulties incurred from the discrete-time analysis.
Additionally,
while an initial attempt to bridge other learning problems to the proposed framework
has been taken in Sec. \ref{sec:beyond-sl},
more are left to be explored.
Specifically,
Generative Adversarial Networks are closely related to
minimax control,
and the dynamical analysis from an SDE viewpoint has been recently discussed to reveal an intriguing variational interpretation
\citep{tao2019variational}.
}

\appendices

\section{}
\label{app:dnn-dynamcs-other-archi}

Critical initialization and mean field approximation can be applied to convolution layers as the number of channels goes to limit \citep{xiao2018dynamical}.
Similar results can be derived except $\bm{\epsilon}_t$ now traverses with a much richer dynamics through convoluted operators.
For recurrent architectures, e.g. RNN and LSTM, the theory suggests that gating mechanisms
facilitate efficient signal propagation \citep{chen2018dynamical,gilboa2019dynamical}.
However, it also casts doubt on several practically-used modules,
as the analysis suggests batch normalization causes exploding gradient signal \citep{schoenholz2016deep}
and dropout destroys the order-to-chaos critical point \citep{yang2019mean}.
Global optimality of GD for other architectures can be proved following similar derivations.
Appendix E in \cite{du2018gradientb} provides a general framework to include FC-DNN, ResNet, and convolution DNN.

\section{}
\label{app:fpe-derivation}
First, we notice that the variational functional $\mathcal{F}_\Psi(\rho; \beta)$ in \eq{\ref{eq:wgf-fpe}} can be written as an
Kullback–Leibler divergence:
\begin{align*}
\mathcal{F}_\Psi(\rho; \beta) := \mathcal{E}_{\Psi}(\rho) - \beta^{-1}\mathcal{S}(\rho) = \beta^{-1}\KL(\rho || \rho^{\mathrm{ss}}) \comma
\end{align*}
where $\rho^{\mathrm{ss}} \propto \exp (-\beta \Psi(x))$.
Now, recall that for a functional $\mathcal{F}: \mathcal{P}_{2} \mapsto \mathbb{R}$ of the following form:
$\mathcal{F}(\rho) := \int_{x} f(\rho(x)) \mathrm{d}x$, its first variation at $\rho$ is given by
$\partial_\rho \mathcal{F}(\rho)(\cdot) = f^{\prime}(\rho(\cdot))$.
Substituting it into \eq{\ref{eq:wgf-fpe}} will lead to \eq{\ref{eq:fpe}}:
\begin{align*}
{\partial_{t} \rho_{t}}
= &\nabla \cdot \left(\rho_{t} \nabla (\partial_\rho \mathcal{F}_\Psi)  \right) \\
= &\beta^{-1}\nabla \cdot \left(\rho_{t} \nabla (\partial_\rho \KL(\rho_t || \rho^{\mathrm{ss}}))  \right) \\
= &\beta^{-1}\nabla \cdot \left(\rho_{t} \nabla \left(
\frac{\mathrm{d}(\rho_t \log \frac{\rho_t}{\rho^{\mathrm{ss}}})}{\mathrm{d}\rho} \right)  \right) \\
= &\beta^{-1}\nabla \cdot \left(\rho_{t} \nabla \left( \log \frac{\rho_t}{\rho^{\mathrm{ss}}} + \cancel{\rho_t}\frac{1}{\cancel{\rho_t}} \right)  \right) \\
= &\beta^{-1}\nabla \cdot \left(\rho_{t} \nabla ( \beta \Psi + \log \rho_t) \right) \\
= &\nabla \cdot \left(\rho_{t} \nabla \Psi\right) + \beta^{-1} \nabla \cdot \left(\rho_{t} \frac{\nabla \rho_t}{\rho_t} \right) \\
= &\nabla \cdot \left(\rho_{t} \nabla \Psi\right) + \beta^{-1} \Delta \rho_{t}
\end{align*}

\section{}
\label{app:master-eq-derive}
Here we recapitulate the derivation from \cite{yaida2018fluctuation}.
First, recall \eq{\ref{eq:discrete-dist-sgd}}:
\begin{align*}
\rho_{t+1}(\vtheta) =& \mathbb{E}_{\thprime \sim \rho_t, \mathcal{B}}   \left[ \deltaa \right] \\
                    =& \mathbb{E}_{\mathcal{B}} \left[ \smallint \dthprime \rho_t(\thprime) \deltaa \right] \period
\end{align*}
The steady-state distribution therefore obeys the relation
\begin{align}
\psss = \expB \left[ \smallint \dthprime \pss(\thprime) \deltaa \right].
\label{eq:ss-dist}
\end{align}
Now, substitute \eq{\ref{eq:ss-dist}} to the expectation of an observable $\mathcal{O}(\vtheta)$ at equilibrium
\begin{equation}
\begin{split}
&\expPss \left[ \mathcal{O}(\vtheta) \right] \\
    =& \smallint \dth \psss \mathcal{O}(\vtheta) \\
    =& \expB \left[ \smallint \dth \smallint \dthprime \pss(\thprime) \deltaa \mathcal{O}(\vtheta) \right] \\
    =& \expB \left[{\smallint \dth} \smallint \dthprime \pss(\thprime) \mathcal{O}(\gmbupdate) \right] \\
    =& \expPss \left[ \expB \left[ \mathcal{O}
    \left(\gmbupdatee\right)
    \right] \right] \period
\end{split}
\end{equation}
We obtain its master equation at equilibrium \eq{\ref{eq:master}}.

\section{}
\label{app:minimax-logsumexp}

Recall the Taylor expansion of $\exp$ and $\log$ functions are
$\exp(x) = 1 + \sum_{k=1}^{\infty} \frac{x^k}{k!}$ , and
$\log(1+x) = \sum_{k=1}^{\infty} (-1)^{k-1} \frac{x^k}{k} $.
Expanding the objective $\mathcal{J}_{k\neq0}$ up to second order leads to
\begin{equation}
\begin{split}
&\quad \mathcal{J}_{k\neq0}(\vx, \xi) \\
=&          \fracK \log \ExpOfXi{ \exp \left( k J \right) } \\
\approx&    \fracK \log \ExpOfXi{ 1 + k J + \frac{k^2}{2} J^2 } \\
\approx&    \fracK      \left[ \left(k \mathbb{E}_\xi J + \frac{k^2}{2} \mathbb{E}_\xi J^2 \right)
                  -\frac{1}{2} \left(k \mathbb{E}_\xi J + \frac{k^2}{2} \mathbb{E}_\xi J^2 \right)^2
                        \right] \quad \\
=&          \fracK      \left[k \mathbb{E}_\xi J + \frac{k^2}{2} \left[ \mathbb{E}_\xi J^2 - (\mathbb{E}_\xi J)^2 \right] + \mathcal{O}(k^{3}) \right] \\
=&          \mathbb{E}_\xi J + \frac{k}{2} \Var_\xi  \left[ J \right] + \mathcal{O}(k^{2}) \period
\end{split}
\end{equation}
For small $k$, the higher-order term $\mathcal{O}(k^{2})$ is negligible and
we obtain the risk-aware interpretation of \eq{\ref{eq:risk-variance-obj}}.

\ifCLASSOPTIONcaptionsoff
  \newpage
\fi

\bibliography{reference}

\begin{thebibliography}{100}
\providecommand{\url}[1]{#1}
\csname url@samestyle\endcsname
\providecommand{\newblock}{\relax}
\providecommand{\bibinfo}[2]{#2}
\providecommand{\BIBentrySTDinterwordspacing}{\spaceskip=0pt\relax}
\providecommand{\BIBentryALTinterwordstretchfactor}{4}
\providecommand{\BIBentryALTinterwordspacing}{\spaceskip=\fontdimen2\font plus
\BIBentryALTinterwordstretchfactor\fontdimen3\font minus
  \fontdimen4\font\relax}
\providecommand{\BIBforeignlanguage}[2]{{%
\expandafter\ifx\csname l@#1\endcsname\relax
\typeout{** WARNING: IEEEtran.bst: No hyphenation pattern has been}%
\typeout{** loaded for the language `#1'. Using the pattern for}%
\typeout{** the default language instead.}%
\else
\language=\csname l@#1\endcsname
\fi
#2}}
\providecommand{\BIBdecl}{\relax}
\BIBdecl

\bibitem{krizhevsky2012imagenet}
A.~Krizhevsky, I.~Sutskever, and G.~E. Hinton, ``Imagenet classification with
  deep convolutional neural networks,'' in \emph{Advances in neural information
  processing systems}, 2012, pp. 1097--1105.

\bibitem{srivastava2012multimodal}
N.~Srivastava and R.~R. Salakhutdinov, ``Multimodal learning with deep
  boltzmann machines,'' in \emph{Advances in neural information processing
  systems}, 2012, pp. 2222--2230.

\bibitem{silver2016mastering}
D.~Silver, A.~Huang, C.~J. Maddison, A.~Guez, L.~Sifre, G.~Van Den~Driessche,
  J.~Schrittwieser, I.~Antonoglou, V.~Panneershelvam, M.~Lanctot \emph{et~al.},
  ``Mastering the game of go with deep neural networks and tree search,''
  \emph{nature}, vol. 529, no. 7587, p. 484, 2016.

\bibitem{geirhos2018imagenet}
R.~Geirhos, P.~Rubisch, C.~Michaelis, M.~Bethge, F.~A. Wichmann, and
  W.~Brendel, ``Imagenet-trained cnns are biased towards texture; increasing
  shape bias improves accuracy and robustness,'' \emph{arXiv preprint
  arXiv:1811.12231}, 2018.

\bibitem{wang2016training}
S.~Wang, W.~Liu, J.~Wu, L.~Cao, Q.~Meng, and P.~J. Kennedy, ``Training deep
  neural networks on imbalanced data sets,'' in \emph{2016 international joint
  conference on neural networks (IJCNN)}.\hskip 1em plus 0.5em minus
  0.4em\relax IEEE, 2016, pp. 4368--4374.

\bibitem{goodfellow2014explaining}
I.~J. Goodfellow, J.~Shlens, and C.~Szegedy, ``Explaining and harnessing
  adversarial examples,'' \emph{arXiv preprint arXiv:1412.6572}, 2014.

\bibitem{athalye2018robustness}
A.~Athalye and N.~Carlini, ``On the robustness of the cvpr 2018 white-box
  adversarial example defenses,'' \emph{arXiv preprint arXiv:1804.03286}, 2018.

\bibitem{simonyan2013deep}
K.~Simonyan, A.~Vedaldi, and A.~Zisserman, ``Deep inside convolutional
  networks: Visualising image classification models and saliency maps,''
  \emph{arXiv preprint arXiv:1312.6034}, 2013.

\bibitem{poole2016exponential}
B.~Poole, S.~Lahiri, M.~Raghu, J.~Sohl-Dickstein, and S.~Ganguli, ``Exponential
  expressivity in deep neural networks through transient chaos,'' in
  \emph{Advances in neural information processing systems}, 2016, pp.
  3360--3368.

\bibitem{dauphin2014identifying}
Y.~N. Dauphin, R.~Pascanu, C.~Gulcehre, K.~Cho, S.~Ganguli, and Y.~Bengio,
  ``Identifying and attacking the saddle point problem in high-dimensional
  non-convex optimization,'' in \emph{Advances in neural information processing
  systems}, 2014, pp. 2933--2941.

\bibitem{zhang2016understanding}
C.~Zhang, S.~Bengio, M.~Hardt, B.~Recht, and O.~Vinyals, ``Understanding deep
  learning requires rethinking generalization,'' \emph{arXiv preprint
  arXiv:1611.03530}, 2016.

\bibitem{sutskever2013importance}
I.~Sutskever, J.~Martens, G.~Dahl, and G.~Hinton, ``On the importance of
  initialization and momentum in deep learning,'' in \emph{International
  conference on machine learning}, 2013, pp. 1139--1147.

\bibitem{hardt2015train}
M.~Hardt, B.~Recht, and Y.~Singer, ``Train faster, generalize better: Stability
  of stochastic gradient descent,'' \emph{arXiv preprint arXiv:1509.01240},
  2015.

\bibitem{chaudhari2018stochastic}
P.~Chaudhari and S.~Soatto, ``Stochastic gradient descent performs variational
  inference, converges to limit cycles for deep networks,'' in \emph{2018
  Information Theory and Applications Workshop (ITA)}.\hskip 1em plus 0.5em
  minus 0.4em\relax IEEE, 2018, pp. 1--10.

\bibitem{keskar2016large}
N.~S. Keskar, D.~Mudigere, J.~Nocedal, M.~Smelyanskiy, and P.~T.~P. Tang, ``On
  large-batch training for deep learning: Generalization gap and sharp
  minima,'' \emph{arXiv preprint arXiv:1609.04836}, 2016.

\bibitem{lee2017first}
J.~D. Lee, I.~Panageas, G.~Piliouras, M.~Simchowitz, M.~I. Jordan, and
  B.~Recht, ``First-order methods almost always avoid saddle points,''
  \emph{arXiv preprint arXiv:1710.07406}, 2017.

\bibitem{lecun1990handwritten}
Y.~LeCun, B.~E. Boser, J.~S. Denker, D.~Henderson, R.~E. Howard, W.~E. Hubbard,
  and L.~D. Jackel, ``Handwritten digit recognition with a back-propagation
  network,'' in \emph{Advances in neural information processing systems}, 1990,
  pp. 396--404.

\bibitem{shwartz2017opening}
R.~Shwartz-Ziv and N.~Tishby, ``Opening the black box of deep neural networks
  via information,'' \emph{arXiv preprint arXiv:1703.00810}, 2017.

\bibitem{du2018gradientb}
S.~S. Du, J.~D. Lee, H.~Li, L.~Wang, and X.~Zhai, ``Gradient descent finds
  global minima of deep neural networks,'' \emph{arXiv preprint
  arXiv:1811.03804}, 2018.

\bibitem{du2018gradienta}
S.~S. Du, X.~Zhai, B.~Poczos, and A.~Singh, ``Gradient descent provably
  optimizes over-parameterized neural networks,'' \emph{arXiv preprint
  arXiv:1810.02054}, 2018.

\bibitem{saxe2018information}
A.~M. Saxe, Y.~Bansal, J.~Dapello, M.~Advani, A.~Kolchinsky, B.~D. Tracey, and
  D.~D. Cox, ``On the information bottleneck theory of deep learning,'' in
  \emph{International Conference on Learning Representations}, 2018.

\bibitem{zou2018stochastic}
D.~Zou, Y.~Cao, D.~Zhou, and Q.~Gu, ``Stochastic gradient descent optimizes
  over-parameterized deep relu networks,'' \emph{arXiv preprint
  arXiv:1811.08888}, 2018.

\bibitem{allen2018convergence}
Z.~Allen-Zhu, Y.~Li, and Z.~Song, ``A convergence theory for deep learning via
  over-parameterization,'' \emph{arXiv preprint arXiv:1811.03962}, 2018.

\bibitem{schoenholz2016deep}
S.~S. Schoenholz, J.~Gilmer, S.~Ganguli, and J.~Sohl-Dickstein, ``Deep
  information propagation,'' \emph{arXiv preprint arXiv:1611.01232}, 2016.

\bibitem{chen2018dynamical}
M.~Chen, J.~Pennington, and S.~S. Schoenholz, ``Dynamical isometry and a mean
  field theory of rnns: Gating enables signal propagation in recurrent neural
  networks,'' \emph{arXiv preprint arXiv:1806.05394}, 2018.

\bibitem{gilboa2019dynamical}
D.~Gilboa, B.~Chang, M.~Chen, G.~Yang, S.~S. Schoenholz, E.~H. Chi, and
  J.~Pennington, ``Dynamical isometry and a mean field theory of lstms and
  grus,'' \emph{arXiv preprint arXiv:1901.08987}, 2019.

\bibitem{xiao2018dynamical}
L.~Xiao, Y.~Bahri, J.~Sohl-Dickstein, S.~S. Schoenholz, and J.~Pennington,
  ``Dynamical isometry and a mean field theory of cnns: How to train
  10,000-layer vanilla convolutional neural networks,'' \emph{arXiv preprint
  arXiv:1806.05393}, 2018.

\bibitem{yang2017mean}
G.~Yang and S.~Schoenholz, ``Mean field residual networks: On the edge of
  chaos,'' in \emph{Advances in neural information processing systems}, 2017,
  pp. 7103--7114.

\bibitem{karakida2018universal}
R.~Karakida, S.~Akaho, and S.-i. Amari, ``Universal statistics of fisher
  information in deep neural networks: Mean field approach,'' \emph{arXiv
  preprint arXiv:1806.01316}, 2018.

\bibitem{pennington2018emergence}
J.~Pennington, S.~S. Schoenholz, and S.~Ganguli, ``The emergence of spectral
  universality in deep networks,'' \emph{arXiv preprint arXiv:1802.09979},
  2018.

\bibitem{williams1997computing}
C.~K. Williams, ``Computing with infinite networks,'' in \emph{Advances in
  neural information processing systems}, 1997, pp. 295--301.

\bibitem{matthews2018gaussian}
A.~G. d.~G. Matthews, M.~Rowland, J.~Hron, R.~E. Turner, and Z.~Ghahramani,
  ``Gaussian process behaviour in wide deep neural networks,'' \emph{arXiv
  preprint arXiv:1804.11271}, 2018.

\bibitem{garriga2018deep}
A.~Garriga-Alonso, L.~Aitchison, and C.~E. Rasmussen, ``Deep convolutional
  networks as shallow gaussian processes,'' \emph{arXiv preprint
  arXiv:1808.05587}, 2018.

\bibitem{lee2017deep}
J.~Lee, Y.~Bahri, R.~Novak, S.~S. Schoenholz, J.~Pennington, and
  J.~Sohl-Dickstein, ``Deep neural networks as gaussian processes,''
  \emph{arXiv preprint arXiv:1711.00165}, 2017.

\bibitem{soudry2018implicit}
D.~Soudry, E.~Hoffer, M.~S. Nacson, S.~Gunasekar, and N.~Srebro, ``The implicit
  bias of gradient descent on separable data,'' \emph{The Journal of Machine
  Learning Research}, vol.~19, no.~1, pp. 2822--2878, 2018.

\bibitem{gunasekar2018characterizing}
S.~Gunasekar, J.~Lee, D.~Soudry, and N.~Srebro, ``Characterizing implicit bias
  in terms of optimization geometry,'' \emph{arXiv preprint arXiv:1802.08246},
  2018.

\bibitem{gunasekar2018implicit}
S.~Gunasekar, J.~D. Lee, D.~Soudry, and N.~Srebro, ``Implicit bias of gradient
  descent on linear convolutional networks,'' in \emph{Advances in Neural
  Information Processing Systems}, 2018, pp. 9482--9491.

\bibitem{ji2018gradient}
Z.~Ji and M.~Telgarsky, ``Gradient descent aligns the layers of deep linear
  networks,'' \emph{arXiv preprint arXiv:1810.02032}, 2018.

\bibitem{allen2018learning}
Z.~Allen-Zhu, Y.~Li, and Y.~Liang, ``Learning and generalization in
  overparameterized neural networks, going beyond two layers,'' \emph{arXiv
  preprint arXiv:1811.04918}, 2018.

\bibitem{wu2018sgd}
L.~Wu, C.~Ma, and E.~Weinan, ``How sgd selects the global minima in
  over-parameterized learning: A dynamical stability perspective,'' in
  \emph{Advances in Neural Information Processing Systems}, 2018, pp.
  8279--8288.

\bibitem{neyshabur2017geometry}
B.~Neyshabur, R.~Tomioka, R.~Salakhutdinov, and N.~Srebro, ``Geometry of
  optimization and implicit regularization in deep learning,'' \emph{arXiv
  preprint arXiv:1705.03071}, 2017.

\bibitem{zhu2018anisotropic}
Z.~Zhu, J.~Wu, B.~Yu, L.~Wu, and J.~Ma, ``The anisotropic noise in stochastic
  gradient descent: Its behavior of escaping from minima and regularization
  effects,'' in \emph{International conference on machine learning}, 2019.

\bibitem{kleinberg2018alternative}
R.~Kleinberg, Y.~Li, and Y.~Yuan, ``An alternative view: When does sgd escape
  local minima?'' \emph{arXiv preprint arXiv:1802.06175}, 2018.

\bibitem{chaudhari2016entropy}
P.~Chaudhari, A.~Choromanska, S.~Soatto, Y.~LeCun, C.~Baldassi, C.~Borgs,
  J.~Chayes, L.~Sagun, and R.~Zecchina, ``Entropy-sgd: Biasing gradient descent
  into wide valleys,'' \emph{arXiv preprint arXiv:1611.01838}, 2016.

\bibitem{chaudhari2018deep}
P.~Chaudhari, A.~Oberman, S.~Osher, S.~Soatto, and G.~Carlier, ``Deep
  relaxation: partial differential equations for optimizing deep neural
  networks,'' \emph{Research in the Mathematical Sciences}, vol.~5, no.~3,
  p.~30, 2018.

\bibitem{teh2016consistency}
Y.~W. Teh, A.~H. Thiery, and S.~J. Vollmer, ``Consistency and fluctuations for
  stochastic gradient langevin dynamics,'' \emph{The Journal of Machine
  Learning Research}, vol.~17, no.~1, pp. 193--225, 2016.

\bibitem{li2016preconditioned}
C.~Li, C.~Chen, D.~Carlson, and L.~Carin, ``Preconditioned stochastic gradient
  langevin dynamics for deep neural networks,'' in \emph{Thirtieth AAAI
  Conference on Artificial Intelligence}, 2016.

\bibitem{tishby2015deep}
N.~Tishby and N.~Zaslavsky, ``Deep learning and the information bottleneck
  principle,'' in \emph{2015 IEEE Information Theory Workshop (ITW)}.\hskip 1em
  plus 0.5em minus 0.4em\relax IEEE, 2015, pp. 1--5.

\bibitem{achille2018critical}
\BIBentryALTinterwordspacing
A.~Achille, M.~Rovere, and S.~Soatto, ``Critical learning periods in deep
  networks,'' in \emph{International Conference on Learning Representations},
  2019. [Online]. Available: \url{https://openreview.net/forum?id=BkeStsCcKQ}
\BIBentrySTDinterwordspacing

\bibitem{achille2018emergence}
A.~Achille and S.~Soatto, ``Emergence of invariance and disentanglement in deep
  representations,'' \emph{The Journal of Machine Learning Research}, vol.~19,
  no.~1, pp. 1947--1980, 2018.

\bibitem{DBLP:journals/corr/abs-1905-12213}
\BIBentryALTinterwordspacing
------, ``Where is the information in a deep neural network?'' \emph{CoRR},
  vol. abs/1905.12213, 2019. [Online]. Available:
  \url{http://arxiv.org/abs/1905.12213}
\BIBentrySTDinterwordspacing

\bibitem{valle-perez2018deep}
\BIBentryALTinterwordspacing
G.~Valle-Perez, C.~Q. Camargo, and A.~A. Louis, ``Deep learning generalizes
  because the parameter-function map is biased towards simple functions,'' in
  \emph{International Conference on Learning Representations}, 2019. [Online].
  Available: \url{https://openreview.net/forum?id=rye4g3AqFm}
\BIBentrySTDinterwordspacing

\bibitem{goldt2017stochastic}
S.~Goldt and U.~Seifert, ``Stochastic thermodynamics of learning,''
  \emph{Physical review letters}, vol. 118, no.~1, p. 010601, 2017.

\bibitem{goldt2017thermodynamic}
------, ``Thermodynamic efficiency of learning a rule in neural networks,''
  \emph{New Journal of Physics}, vol.~19, no.~11, p. 113001, 2017.

\bibitem{yaida2018fluctuation}
S.~Yaida, ``Fluctuation-dissipation relations for stochastic gradient
  descent,'' \emph{arXiv preprint arXiv:1810.00004}, 2018.

\bibitem{weinan2017proposal}
E.~Weinan, ``A proposal on machine learning via dynamical systems,''
  \emph{Communications in Mathematics and Statistics}, vol.~5, no.~1, pp.
  1--11, 2017.

\bibitem{he2016deep}
K.~He, X.~Zhang, S.~Ren, and J.~Sun, ``Deep residual learning for image
  recognition,'' in \emph{Proceedings of the IEEE conference on computer vision
  and pattern recognition}, 2016, pp. 770--778.

\bibitem{lu2017beyond}
Y.~Lu, A.~Zhong, Q.~Li, and B.~Dong, ``Beyond finite layer neural networks:
  Bridging deep architectures and numerical differential equations,''
  \emph{arXiv preprint arXiv:1710.10121}, 2017.

\bibitem{JMLR:v20:16-243}
\BIBentryALTinterwordspacing
S.~Sonoda and N.~Murata, ``Transport analysis of infinitely deep neural
  network,'' \emph{Journal of Machine Learning Research}, vol.~20, no.~2, pp.
  1--52, 2019. [Online]. Available:
  \url{http://jmlr.org/papers/v20/16-243.html}
\BIBentrySTDinterwordspacing

\bibitem{chen2018neural}
T.~Q. Chen, Y.~Rubanova, J.~Bettencourt, and D.~K. Duvenaud, ``Neural ordinary
  differential equations,'' in \emph{Advances in Neural Information Processing
  Systems}, 2018, pp. 6572--6583.

\bibitem{chen2019cheapdiffopt}
R.~T.~Q. Chen and D.~Duvenaud, ``Neural networks with cheap differential
  operators,'' in \emph{2019 ICML Workshop on Invertible Neural Nets and
  Normalizing Flows (INNF)}, 2019.

\bibitem{hu2017control}
B.~Hu and L.~Lessard, ``Control interpretations for first-order optimization
  methods,'' in \emph{2017 American Control Conference (ACC)}.\hskip 1em plus
  0.5em minus 0.4em\relax IEEE, 2017, pp. 3114--3119.

\bibitem{han2018mean}
E.~Weinan, J.~Han, and Q.~Li, ``A mean-field optimal control formulation of
  deep learning,'' \emph{arXiv preprint arXiv:1807.01083}, 2018.

\bibitem{li2017maximum}
Q.~Li, L.~Chen, C.~Tai, and E.~Weinan, ``Maximum principle based algorithms for
  deep learning,'' \emph{The Journal of Machine Learning Research}, vol.~18,
  no.~1, pp. 5998--6026, 2017.

\bibitem{li2018optimal}
Q.~Li and S.~Hao, ``An optimal control approach to deep learning and
  applications to discrete-weight neural networks,'' \emph{arXiv preprint
  arXiv:1803.01299}, 2018.

\bibitem{zhang2019you}
D.~Zhang, T.~Zhang, Y.~Lu, Z.~Zhu, and B.~Dong, ``You only propagate once:
  Accelerating adversarial training via maximal principle,'' \emph{arXiv
  preprint arXiv:1905.00877}, 2019.

\bibitem{pavliotis2014stochastic}
G.~A. Pavliotis, \emph{Stochastic processes and applications: diffusion
  processes, the Fokker-Planck and Langevin equations}.\hskip 1em plus 0.5em
  minus 0.4em\relax Springer, 2014, vol.~60.

\bibitem{simsekli2019tail}
U.~Simsekli, L.~Sagun, and M.~Gurbuzbalaban, ``A tail-index analysis of
  stochastic gradient noise in deep neural networks,'' \emph{arXiv preprint
  arXiv:1901.06053}, 2019.

\bibitem{li2017stochastic}
Q.~Li, C.~Tai, and W.~E, ``Stochastic modified equations and adaptive
  stochastic gradient algorithms,'' in \emph{Proceedings of the 34th
  International Conference on Machine Learning-Volume 70}.\hskip 1em plus 0.5em
  minus 0.4em\relax JMLR. org, 2017, pp. 2101--2110.

\bibitem{an2018stochastic}
J.~An, J.~Lu, and L.~Ying, ``Stochastic modified equations for the asynchronous
  stochastic gradient descent,'' \emph{arXiv preprint arXiv:1805.08244}, 2018.

\bibitem{glorot2010understanding}
X.~Glorot and Y.~Bengio, ``Understanding the difficulty of training deep
  feedforward neural networks,'' in \emph{Proceedings of the thirteenth
  international conference on artificial intelligence and statistics}, 2010,
  pp. 249--256.

\bibitem{li2018learning}
Y.~Li and Y.~Liang, ``Learning overparameterized neural networks via stochastic
  gradient descent on structured data,'' in \emph{Advances in Neural
  Information Processing Systems}, 2018, pp. 8157--8166.

\bibitem{jacot2018neural}
A.~Jacot, F.~Gabriel, and C.~Hongler, ``Neural tangent kernel: Convergence and
  generalization in neural networks,'' in \emph{Advances in neural information
  processing systems}, 2018, pp. 8571--8580.

\bibitem{qian1999momentum}
N.~Qian, ``On the momentum term in gradient descent learning algorithms,''
  \emph{Neural networks}, vol.~12, no.~1, pp. 145--151, 1999.

\bibitem{nokland2019training}
A.~N{\o}kland and L.~H. Eidnes, ``Training neural networks with local error
  signals,'' \emph{arXiv preprint arXiv:1901.06656}, 2019.

\bibitem{jaderberg2016reinforcement}
M.~Jaderberg, V.~Mnih, W.~M. Czarnecki, T.~Schaul, J.~Z. Leibo, D.~Silver, and
  K.~Kavukcuoglu, ``Reinforcement learning with unsupervised auxiliary tasks,''
  \emph{arXiv preprint arXiv:1611.05397}, 2016.

\bibitem{liu2017learning}
G.-H. Liu, A.~Siravuru, S.~Prabhakar, M.~Veloso, and G.~Kantor, ``Learning
  end-to-end multimodal sensor policies for autonomous navigation,''
  \emph{arXiv preprint arXiv:1705.10422}, 2017.

\bibitem{boltyanskii1960theory}
V.~G. Boltyanskii, R.~V. Gamkrelidze, and L.~S. Pontryagin, ``The theory of
  optimal processes. i. the maximum principle,'' TRW SPACE TECHNOLOGY LABS LOS
  ANGELES CALIF, Tech. Rep., 1960.

\bibitem{bertsekas1995dynamic}
D.~P. Bertsekas, D.~P. Bertsekas, D.~P. Bertsekas, and D.~P. Bertsekas,
  \emph{Dynamic programming and optimal control}.\hskip 1em plus 0.5em minus
  0.4em\relax Athena scientific Belmont, MA, 1995, vol.~1, no.~2.

\bibitem{bellman1964selected}
R.~E. Bellman and R.~E. Kalaba, \emph{Selected papers on mathematical trends in
  control theory}.\hskip 1em plus 0.5em minus 0.4em\relax Dover Publications,
  1964.

\bibitem{ambrosio2008gradient}
L.~Ambrosio, N.~Gigli, and G.~Savar{\'e}, \emph{Gradient flows: in metric
  spaces and in the space of probability measures}.\hskip 1em plus 0.5em minus
  0.4em\relax Springer Science \& Business Media, 2008.

\bibitem{franklin1994feedback}
G.~F. Franklin, J.~D. Powell, A.~Emami-Naeini, and J.~D. Powell, \emph{Feedback
  control of dynamic systems}.\hskip 1em plus 0.5em minus 0.4em\relax
  Addison-Wesley Reading, MA, 1994, vol.~3.

\bibitem{rotskoff2019neural}
G.~M. Rotskoff and E.~Vanden-Eijnden, ``Trainability and accuracy of neural
  networks: An interacting particle system approach,'' \emph{arXiv preprint
  arXiv:1805.00915v3}, 2019.

\bibitem{oksendal2003stochastic}
B.~{\O}ksendal, ``Stochastic differential equations,'' in \emph{Stochastic
  differential equations}.\hskip 1em plus 0.5em minus 0.4em\relax Springer,
  2003, pp. 65--84.

\bibitem{smith2017bayesian}
S.~L. Smith and Q.~V. Le, ``A bayesian perspective on generalization and
  stochastic gradient descent,'' \emph{arXiv preprint arXiv:1710.06451}, 2017.

\bibitem{jastrzkebski2017three}
S.~Jastrzebski, Z.~Kenton, D.~Arpit, N.~Ballas, A.~Fischer, Y.~Bengio, and
  A.~Storkey, ``Three factors influencing minima in sgd,'' \emph{arXiv preprint
  arXiv:1711.04623}, 2017.

\bibitem{moulines2011non}
E.~Moulines and F.~R. Bach, ``Non-asymptotic analysis of stochastic
  approximation algorithms for machine learning,'' in \emph{Advances in Neural
  Information Processing Systems}, 2011, pp. 451--459.

\bibitem{milstein1994numerical}
G.~N. Milstein, \emph{Numerical integration of stochastic differential
  equations}.\hskip 1em plus 0.5em minus 0.4em\relax Springer Science \&
  Business Media, 1994, vol. 313.

\bibitem{ito1951stochastic}
K.~It{\^o}, \emph{On stochastic differential equations}.\hskip 1em plus 0.5em
  minus 0.4em\relax American Mathematical Soc., 1951, vol.~4.

\bibitem{kolmogoroff1931analytischen}
A.~Kolmogoroff, ``{\"U}ber die analytischen methoden in der
  wahrscheinlichkeitsrechnung,'' \emph{Mathematische Annalen}, vol. 104, no.~1,
  pp. 415--458, 1931.

\bibitem{jordan1998variational}
R.~Jordan, D.~Kinderlehrer, and F.~Otto, ``The variational formulation of the
  fokker--planck equation,'' \emph{SIAM journal on mathematical analysis},
  vol.~29, no.~1, pp. 1--17, 1998.

\bibitem{bovier2004metastability}
A.~Bovier, M.~Eckhoff, V.~Gayrard, and M.~Klein, ``Metastability in reversible
  diffusion processes i: Sharp asymptotics for capacities and exit times,''
  \emph{Journal of the European Mathematical Society}, vol.~6, no.~4, pp.
  399--424, 2004.

\bibitem{hu2017diffusion}
W.~Hu, C.~J. Li, L.~Li, and J.-G. Liu, ``On the diffusion approximation of
  nonconvex stochastic gradient descent,'' \emph{arXiv preprint
  arXiv:1705.07562}, 2017.

\bibitem{xu2011towards}
W.~Xu, ``Towards optimal one pass large scale learning with averaged stochastic
  gradient descent,'' \emph{arXiv preprint arXiv:1107.2490}, 2011.

\bibitem{coraluppi1999risk}
S.~P. Coraluppi and S.~I. Marcus, ``Risk-sensitive and minimax control of
  discrete-time, finite-state markov decision processes,'' \emph{Automatica},
  vol.~35, no.~2, pp. 301--309, 1999.

\bibitem{bacsar2008h}
T.~Ba{\c{s}}ar and P.~Bernhard, \emph{H-infinity optimal control and related
  minimax design problems: a dynamic game approach}.\hskip 1em plus 0.5em minus
  0.4em\relax Springer Science \& Business Media, 2008.

\bibitem{welling2011bayesian}
M.~Welling and Y.~W. Teh, ``Bayesian learning via stochastic gradient langevin
  dynamics,'' in \emph{Proceedings of the 28th international conference on
  machine learning (ICML-11)}, 2011, pp. 681--688.

\bibitem{szegedy2013intriguing}
C.~Szegedy, W.~Zaremba, I.~Sutskever, J.~Bruna, D.~Erhan, I.~Goodfellow, and
  R.~Fergus, ``Intriguing properties of neural networks,'' \emph{arXiv preprint
  arXiv:1312.6199}, 2013.

\bibitem{goodfellow2018defense}
I.~Goodfellow, ``Defense against the dark arts: An overview of adversarial
  example security research and future research directions,'' \emph{arXiv
  preprint arXiv:1806.04169}, 2018.

\bibitem{schmidt2018adversarially}
L.~Schmidt, S.~Santurkar, D.~Tsipras, K.~Talwar, and A.~Madry, ``Adversarially
  robust generalization requires more data,'' in \emph{Advances in Neural
  Information Processing Systems}, 2018, pp. 5014--5026.

\bibitem{wong2017provable}
E.~Wong and J.~Z. Kolter, ``Provable defenses against adversarial examples via
  the convex outer adversarial polytope,'' \emph{arXiv preprint
  arXiv:1711.00851}, 2017.

\bibitem{finn2017model}
C.~Finn, P.~Abbeel, and S.~Levine, ``Model-agnostic meta-learning for fast
  adaptation of deep networks,'' in \emph{Proceedings of the 34th International
  Conference on Machine Learning-Volume 70}.\hskip 1em plus 0.5em minus
  0.4em\relax JMLR. org, 2017, pp. 1126--1135.

\bibitem{duan2016rl}
Y.~Duan, J.~Schulman, X.~Chen, P.~L. Bartlett, I.~Sutskever, and P.~Abbeel,
  ``Rl{$^2$}: Fast reinforcement learning via slow reinforcement learning,''
  \emph{arXiv preprint arXiv:1611.02779}, 2016.

\bibitem{grant2018recasting}
E.~Grant, C.~Finn, S.~Levine, T.~Darrell, and T.~Griffiths, ``Recasting
  gradient-based meta-learning as hierarchical bayes,'' \emph{arXiv preprint
  arXiv:1801.08930}, 2018.

\bibitem{tao2019variational}
C.~Tao, S.~Dai, L.~Chen, K.~Bai, J.~Chen, C.~Liu, R.~Zhang, G.~Bobashev, and
  L.~C. Duke, ``Variational annealing of gans: A langevin perspective,'' in
  \emph{International Conference on Machine Learning}, 2019, pp. 6176--6185.

\bibitem{yang2019mean}
G.~Yang, J.~Pennington, V.~Rao, J.~Sohl-Dickstein, and S.~S. Schoenholz, ``A
  mean field theory of batch normalization,'' \emph{arXiv preprint
  arXiv:1902.08129}, 2019.

\end{thebibliography}

\bibliographystyle{IEEEtran}





\end{document}